\pdfoutput=1
\documentclass[11pt]{article}
\usepackage{authblk}

\usepackage[]{EMNLP2023}

\usepackage{times}
\usepackage{latexsym}

\usepackage[T2A,T1]{fontenc}
\usepackage[utf8]{inputenc}
\usepackage[hindi,russian,english]{babel}
\usepackage{microtype}
\usepackage{amsmath}
\usepackage{graphicx}
\usepackage{subfigure}
\usepackage{amsfonts}
\usepackage{multirow}
\usepackage{listings}
\usepackage{url}
\usepackage[ruled,linesnumbered]{algorithm2e}  
\usepackage{CJKutf8}
\usepackage{graphicx}
\usepackage{subfigure}
\usepackage{supertabular}
\usepackage{subfigure}
\usepackage{dcolumn,booktabs}
\usepackage{hyperref}
\usepackage{tikz}
\usepackage[right]{rotating}

\usepackage{inconsolata}

\title{\frameworkname: A Framework of Initializing Unseen Subword Embeddings for \\ Efficient Large-scale Multilingual Continued Pretraining}

\author[*$\diamond$]{\bf Yihong Liu}
\author[*$\diamond$]{\bf Peiqin Lin}
\author[*$\dagger$]{\bf Mingyang Wang}
\author[*$\diamond$]{\bf Hinrich Sch\"utze}

\affil[*]{Center for Information and Language Processing, LMU Munich} \affil[$\diamond$]{Munich Center for Machine Learning (MCML)} \affil[$\dagger$]{Bosch Center for Artificial Intelligence
 \protect\\ \texttt{\{yihong, linpq, mingyang\}@cis.lmu.de}}

\def\networktwo{ColexNet+\xspace}

\def\embname{$\overrightarrow{\mbox{\networktwo}}$\xspace}
\def\frameworkname{\textsc{Ofa}\xspace}

\def\wechsel{\textsc{Wechsel}\xspace}

\definecolor{green}{RGB}{197,224,180}
\definecolor{blue}{RGB}{143,170,220}
\definecolor{orange}{RGB}{224,177,131}

\def\secref#1{\S\ref{sec:#1}}
\def\seclabel#1{\label{sec:#1}}

\newcounter{notecounter}
\newcommand{\enotesoff}{\long\gdef\enote##1##2{}}
\newcommand{\enoteson}{\long\gdef\enote##1##2{{
\stepcounter{notecounter}
{\large\bf \hspace{1cm}\arabic{notecounter} $<<<$ ##1: ##2 $>>>$\hspace{1cm}}}}}
\enoteson
 \enotesoff
 
\begin{document}
\maketitle
\begin{abstract}

% Pretraining multilingual language models from scratch requires considerable computational resources and substantial training data.
% Therefore, 
Instead of pretraining multilingual language models from scratch, a more efficient method is to adapt existing pretrained language models (PLMs) to new languages via vocabulary 
extension and continued pretraining. 
However, this method usually randomly initializes the embeddings of new subwords and introduces substantially more embedding parameters to the model, thus weakening the efficiency.
To address these issues, 
we propose a novel framework: \textbf{O}ne \textbf{F}or \textbf{A}ll (\textbf{\textsc{Ofa}}), which wisely initializes the embeddings of unseen subwords and thus can adapt a PLM to multiple languages efficiently and effectively.
\textsc{Ofa} takes advantage of external well-aligned multilingual static word vectors and injects the alignment knowledge into the subword embeddings. In addition, \textsc{Ofa} 
applies matrix factorization and replaces the cumbersome embeddings with two lower-dimensional matrices, 
which largely reduces the number of parameters. 
We show \textsc{Ofa} accelerates the convergence of continued pretraining, 
which is environmentally friendly as much fewer carbon footprints are generated. 
Through extensive experiments, we demonstrate \textsc{Ofa} can achieve competitive or better performance than default continued pretraining baselines on a wide range of crosslingual downstream tasks. We make our code and models publicly available.\footnote{\url{https://github.com/cisnlp/ofa}}
\end{abstract}

\section{Introduction}

\begin{figure}[htbp]
    \setlength{\belowcaptionskip}{-0.6cm}
    \centering
    \subfigure[RoBERTa-based models]{
        \includegraphics[width=1.42in]{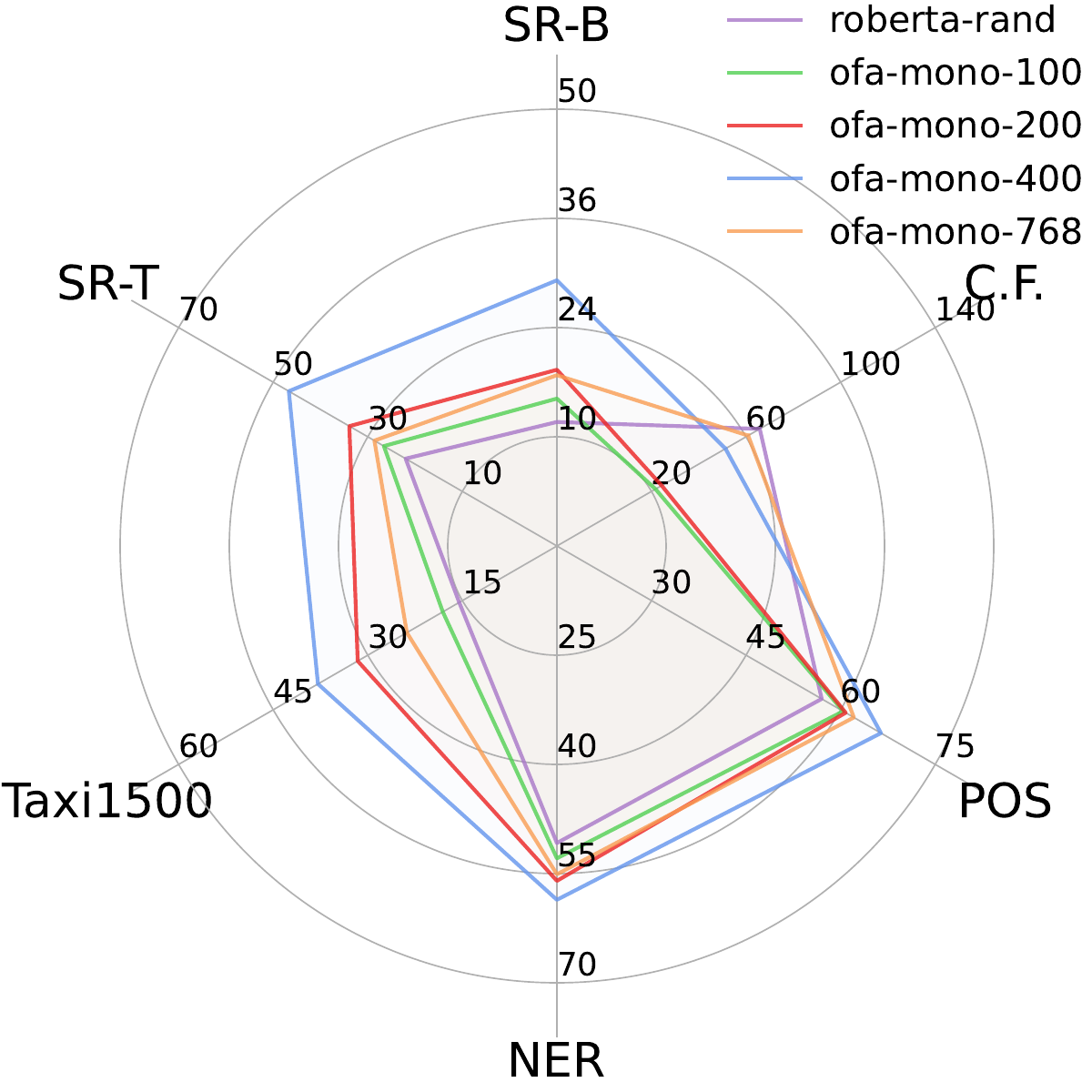}
    }
    \subfigure[XLM-R-based models]{
	\includegraphics[width=1.42in]{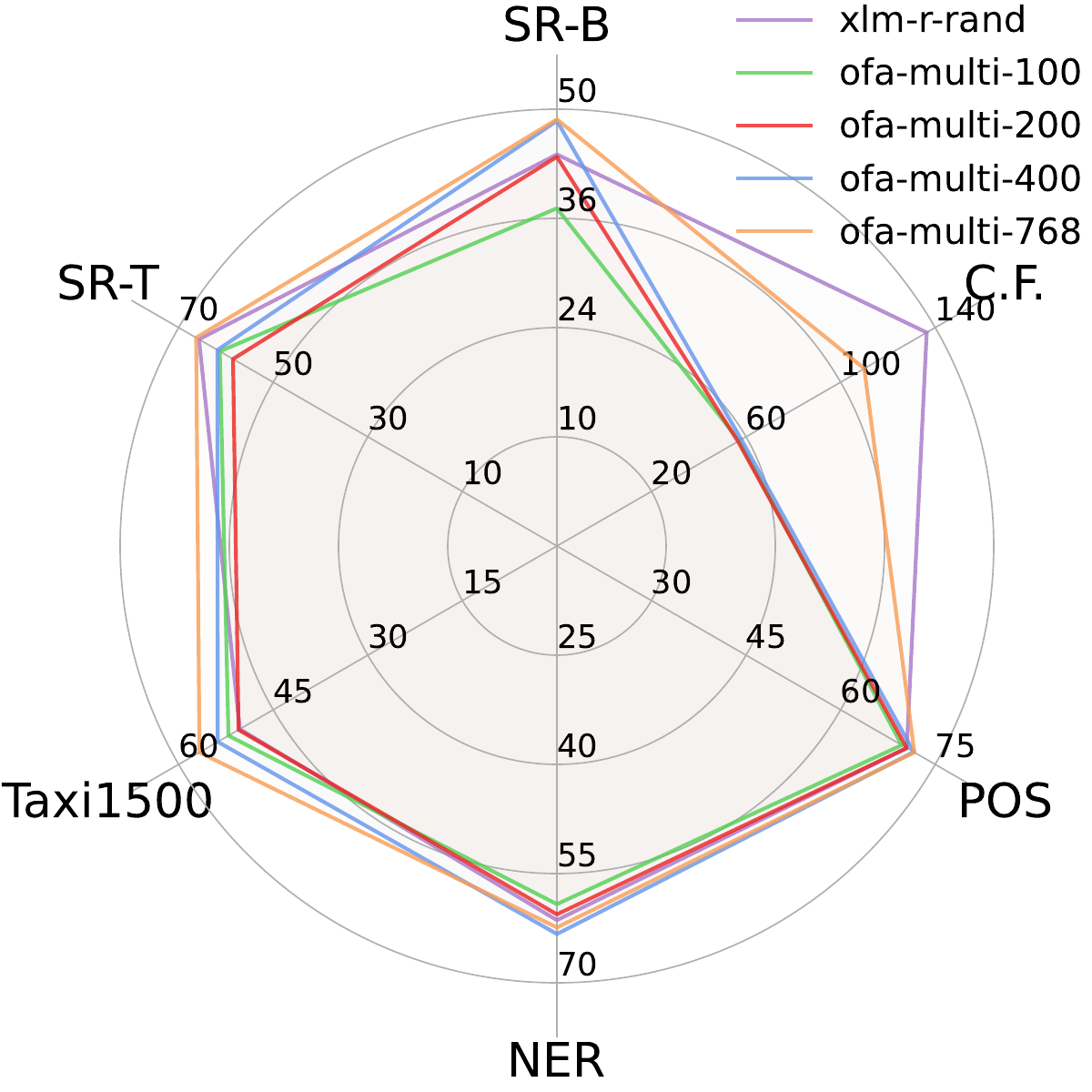}
    }
    \caption{Qualitative comparisons between baselines and \frameworkname. \frameworkname consistently achieves competitive or better performance than the baselines using both (a) monolingual (RoBERTa) or (b) multilingual (XLM-R) PLMs as the source model, with fewer carbon footprints (C.F.) during the continued pretraining, indicating higher efficiency. The stride of each axis in the chart is different.}
    \label{fig:first_page}
\end{figure}

Multilingual PLMs (mPLMs), such as mBERT \citep{devlin-etal-2019-bert} and XLM-R \citep{conneau-etal-2020-unsupervised}, have demonstrated remarkable zero-shot crosslingual capability \citep{huang-etal-2019-unicoder,artetxe-etal-2020-cross}. That is, with only finetuning in some (high-resource) languages to perform a task, the multilingual model can be directly applied to other (low-resource) languages. However, training such mPLMs from scratch requires massive data of different languages, and most importantly, considerable computing resources and energy \citep{wang-etal-2019-improving,bender2021dangers,zhou2023comprehensive}. Therefore, continued pretraining from existing models has been a good alternative \citep{wang-etal-2022-expanding,alabi-etal-2022-adapting,imani-etal-2023-glot500}. However, two problems are generally overlooked in the context of multilingual continued pertaining with vocabulary extension: \textbf{(a)} the random initialization of embeddings for new subwords does not actively use any lexical knowledge encoded in the model; \textbf{(b)} the introduction of many new parameters may pose efficiency problem.

Regarding \textbf{(a)}, the default random initialization approach which samples from a given distribution, e.g., a Gaussian
% with a mean and variance of the original embedding matrix
\citep{hewitt2021initializing,de-vries-nissim-2021-good,marchisio-etal-2023-mini}, does not actively use the lexical knowledge of the original embeddings. To better leverage existing knowledge, some recent works propose to initialize the embeddings for target-language subwords by exploiting both external crosslingual static word vectors and the original PLM embeddings \citep{tran2020english, minixhofer-etal-2022-wechsel, dobler2023focus}. 
Unfortunately, these methods either bilingualize a PLM or create a new monolingual LM for a single target language at a time, which is not ideal in the context of multilingual continued pretraining. Therefore, our goal is to adapt to many languages all at once and wisely initialize the new subword embeddings for large-scale multilingual continued pretraining.

Regarding \textbf{(b)}, adapting to more languages will unarguably introduce more parameters. According to \citet{Chung2021rethinking}, the embedding matrix of multilingual models makes up around 50\% of the model's entire parameters. This percentage can be further increased when adding more new subwords as a consequence of adapting to more languages. 
For example, XLM-V \citep{liang-etal-2023-xlm} increases its vocabulary to 901K,
resulting in embeddings occupying approximately 90\% of the overall parameters.
In the monolingual setting, the factorized embedding parameterization shows effectiveness without sacrificing much performance \citep{lan2020albert}. Thus, a similar method is expected to succeed in multilingual models, given that embeddings are inherently more redundant: \emph{words from different languages that refer to the same concept often have similar representations}. Therefore, we aim to reduce the number of parameters in the embeddings through factorized parameterization.

To this end, we introduce \textbf{\frameworkname}, a framework that wisely initializes the embeddings of new subwords with a factorized parameterization for efficient large-scale multilingual continued pretraining. 
\frameworkname first factorizes the embeddings of the source PLM and uses two smaller matrices to replace it. In the lower-dimensional space, the embeddings of the non-shared new subwords are represented as combinations of the embeddings of some subwords from the source PLM, weighted by the similarity extracted from well-aligned external static multilingual vectors \citep{liu2023crosslingual} that cover 1,335 languages. The embeddings of the shared subwords are directly copied. Finally, \frameworkname copies all non-embedding parameters of the source PLM model and exchanges the source tokenizer (the tokenizer of the source PLM) with the target tokenizer (the tokenizer after vocabulary extension). 

We use a monolingual PLM, i.e., RoBERTa \citep{liu2019roberta} and a multilingual PLM, i.e., XLM-R \citep{conneau-etal-2020-unsupervised} as our source models.
We first apply \frameworkname to these models and then continued pretrain the resulting models on the Glot500-c corpus \citep{imani-etal-2023-glot500}. The final models are evaluated on a diverse set of downstream tasks, including sentence retrieval, text classification, and sequence labeling. \frameworkname not only accelerates the convergence of continued pretraining thus much fewer carbon footprints are generated, but also achieves competitive or better performance on all tasks compared with randomly initialized or full-dimensional baselines, as shown in Figure \ref{fig:first_page}. 

The contributions of this work are as follows: (i) We propose \frameworkname, a framework that wisely initializes the embeddings of unseen subwords with factorized parametrization, targeted on efficient multilingual continued pretraining. (ii) We conduct extensive and strictly controlled experiments on a wide range of downstream tasks and show that \frameworkname is effective and boosts crosslingual transfer. (iii) We show \frameworkname is efficient and environmentally friendly: achieving better performance with less GPU consumption and fewer carbon footprints.

\section{Related Work}
There are generally two ways to obtain a multilingual PLM.
The first way is to pretrain a model from scratch directly on a number of languages with a specific self-learning objective, e.g., masked language modeling (MLM) \citep{devlin-etal-2019-bert}. The typical models that adopt such a strategy are encoder-only models such as mBERT \citep{devlin-etal-2019-bert}, XLM-R \citep{conneau-etal-2020-unsupervised}, IndicBERT \citep{kakwani-etal-2020-indicnlpsuite}, AfriBERTa \citep{ogueji-etal-2021-small} and XLM-V \citep{liang-etal-2023-xlm}, decoder-only models such as XGLM \citep{lin-etal-2022-shot}, mGPT \citep{shliazhko2022mgpt} and BLOOM \citep{scao2022bloom}, and encoder-decoder models such as mBART \citep{liu-etal-2020-multilingual-denoising} and mT5 \citep{xue-etal-2021-mt5}. The alternative way is to use publicly available multilingual PLMs as the source models and continued pretrain them on a set of target languages \citep{wang-etal-2022-expanding,alabi-etal-2022-adapting,imani-etal-2023-glot500}. This continued pretraining approach is in favor because it consumes fewer resources than training from scratch, which is important when the computation budget is limited given the continually increasing model size \citep{tay2022sacle,gupta2023continual}.

One key reason why this continued pretraining approach works is the crosslingual ability of the original multilingual PLMs \citep{pires-etal-2019-multilingual,k2020crosslingual,chai-etal-2022-cross}. With this ability, during continued pretraining, the model could leverage the knowledge gained in the previous pretraining phase as a prior, and adapt to the new languages quickly. 
Some prior works attempt to actively capitalize latent
knowledge encoded in the parameters (embeddings or the
transformer body) of the source
PLM \citep{artetxe-etal-2020-cross,pfeiffer-etal-2021-unks}
when transferring to new languages. However, embeddings of
new subwords are randomly initialized. Most
recently, \citet{tran2020english}, \citet{minixhofer-etal-2022-wechsel}
and \citet{dobler2023focus} explore the possibility of
leveraging both the source PLM embeddings and well-aligned
external crosslingual word vectors to initialize the
embeddings of new subwords for a \textbf{single} target
language at a time. However, how this type of method could be efficiently applied to multilingual scenarios is left unexplored. Our work, in contrast to former research, aims to establish a framework to adapt a PLM, regardless of monolingual or multilingual, to multiple languages. In addition, our framework is targeted towards parameter efficiency, which is friendly to a limited computation budget. 

Our work is also related to some approaches that try to extend the vocabulary of a PLM for specific downstream tasks \citep{wang-etal-2019-improving,tai-etal-2020-exbert,hong-etal-2021-avocado,nag-etal-2023-entropy}. This line of work usually learns the additive vocabulary from the new domain data and therefore specializes a PLM to certain domains. In contrast, our work aims to build a framework to strengthen the crosslinguality of an mPLM for general purposes instead of focusing on specific downstream tasks. This is achieved partly by using the external multilingual word vectors from which some alignment knowledge could be injected into the newly initialized subword embeddings. In this perspective, our work is also related to some post-pretraining alignment methods \citep{pan-etal-2021-multilingual,feng-etal-2022-language,ji-etal-2023-isotropic,liu2024translico} that use word correspondence, translation or transliteration to improve the crosslingual transfer ability of mPLMs.

\section{Preliminary: Embedding Factorization}
% matrix factorization on the embeddings

Before stepping into \frameworkname framework, we first introduce one key technique used by \frameworkname: source embedding factorization. Although matrix factorization itself is not new and is widely leveraged, e.g., in ALBERT \citep{lan2020albert} (a monolingual model) to lower memory consumption. We instead look at this factorization from a \textbf{multilingual perspective} and provide the intuition as to why such low-rank parameterization is effective in large-scale \textbf{multilingual continued pretraining}. 

Given the embeddings $\boldsymbol{E}^{s} \in \mathbb{R}^{|V^s| \times D}$ from a source PLM that is pretrained on some source languages $S$,
% ($S$ can be a single language; or group of languages: a multilingual PLM)
where $V^s$ is its subword vocabulary and $D$ is the embedding dimension, we propose to factorize the matrix $\boldsymbol{E}^{s}$ into lower-dimensional embeddings $\boldsymbol{F}^{s} \in \mathbb{R}^{|V^s| \times D^\prime}$ and an orthogonal up-projection matrix $\boldsymbol{P} \in \mathbb{R}^{D^\prime \times D}$: $
    \boldsymbol{E}^{s} \approx \boldsymbol{F}^{s} \boldsymbol{P}
$, where 
$D^\prime < D$.
$\boldsymbol{P}$ can be interpreted as the embeddings of
a set of $D^\prime$ latent semantic concepts that are language-agnostic. Each concept is represented as a $D$-dimensional vector and these vectors as a whole serve as the basis of a semantic space in $\mathbb{R}^D$ (a subspace in $\mathbb{R}^D$ as $D^\prime < D$) for all subwords. Thus we refer to $\boldsymbol{P}$ as the \emph{primitive embeddings}. $\boldsymbol{F}^{s}$ can be regarded as \emph{coordinates} of all subwords in $V^s$ in the space spanned by $\boldsymbol{P}$. The final representation of a subword $v$ will be the linear combination of the primitive embeddings according to the corresponding coordinates $\boldsymbol{F}^{s}_{\{v\}}$: $\boldsymbol{P}^T \boldsymbol{F}^{s}_{\{v\}}$. 

By factorizing the embeddings into the language-agnostic part $\boldsymbol{P}$ and language-specific part $\boldsymbol{F}^{s}$, we can reduce the number of trainable parameters from $|V^s| \times D$ to $|V^s| \times D^\prime + D^\prime \times D$. This reduction of parameters can be prominent when $D^\prime \ll D$. In addition, as $\boldsymbol{P}$ is shared across languages, we only need to find the target coordinates $\boldsymbol{F}^{t} \in \mathbb{R}^{|V^t| \times D^\prime}$ under the same basis $\boldsymbol{P}$ when we want to adapt the model to new languages whose vocabulary is $V^t$. This is much more efficient than finding $\boldsymbol{E}^{t} \in \mathbb{R}^{|V^t| \times D}$, considering $|V^t|$ can be considerably large in a multilingual setting. Lastly, any coordinates in $\boldsymbol{F}^{t}$ can be up-projected back to $\mathbb{R}^D$ through $\boldsymbol{P}$, corresponding to the hidden size of the transformer body of the source PLM.

\begin{figure}
    \setlength{\belowcaptionskip}{-0.4cm}
  \centering
  \includegraphics[width=0.48\textwidth]{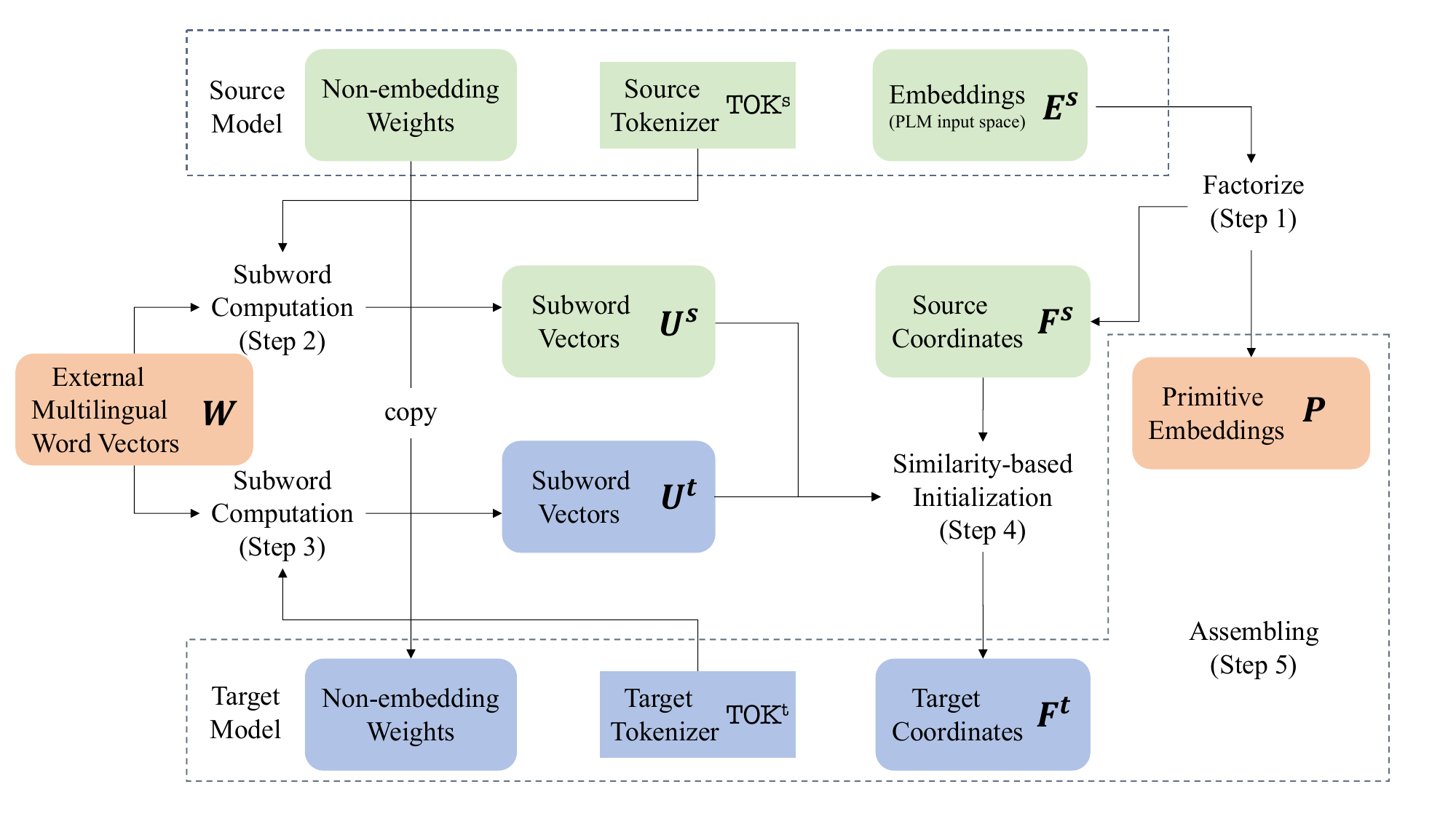}
  \caption{Summary of \frameworkname. Different color indicates the block is specific to different languages. \textcolor{green}{Green}: source languages; \textcolor{blue}{blue}: target languages; \textcolor{orange}{orange}: both.}
  \label{fig:framework}
\end{figure}

\section{\frameworkname Framework}

\frameworkname initializes the embeddings of new subwords in a factorized parametrization. The basic idea of \frameworkname is as follows. We leverage an external multilingual word vector\footnote{To avoid confusion, we use the word ``word vectors'' to refer to any vector in the external static word vector space, and ``embedding'' to refer to the embeddings in the PLM space.} space (which provides high-quality representations of both source and target languages) to induce a measure of semantic similarity on the joint set of subwords and words of both source and target languages. This similarity measure then allows us to initialize subwords of target languages with semantically meaningful representations in the source PLM embedding space.
% The resulting models serve as good starting points for efficient multilingual continued preratining.
We show the summary of \frameworkname framework in Figure \ref{fig:framework} and describe the process step by step as follows.

\paragraph{Problem Setting.} Given well-aligned external
static multilingual word vectors $\boldsymbol{W}$ (vocabulary
$V$), a source PLM (subword embeddings are
$\boldsymbol{E}^{s}$) with its tokenizer $\texttt{TOK}^s$
(vocabulary $V^s$) and target tokenizer $\texttt{TOK}^t$ (vocabulary $V^t$), we want to find \textbf{a good initialization} of embeddings for all subwords in $V^t$, i.e., $\boldsymbol{F}^{t}$, which are \textbf{in lower dimensions}.

\paragraph{Step 1.} We factorize $\boldsymbol{E}^{s}$ from the source PLM to primitive embeddings $\boldsymbol{P}$ and source coordinates $\boldsymbol{F}^{s}$. $\boldsymbol{P}$ will serve as the base of subword embeddings for all languages, and $\boldsymbol{F}^{s}$ will be used to initialize the desired target coordinates $\boldsymbol{F}^{t}$ in \textbf{Step 4}. We simply let $\boldsymbol{F}^{s} = \boldsymbol{E}^{s}$ for baseline models where no matrix factorization is applied to $\boldsymbol{E}^{s}$.

\paragraph{Step 2.} We use the source tokenizer $\texttt{TOK}^s$ to tokenize all words in $V$.
% i.e., the vocabulary of the external multilingual word vectors $\boldsymbol{W}$
We then create a directed bipartite graph between words in $V$ and subwords in $V^s$ that can be tokenized from those words.  We use \embname \citep{liu2023crosslingual} as the word vectors, as they show very strong crosslinguality and reflect conceptual similarity \citep{liu-etal-2023-crosslingual,ye2023study} in many languages (see \secref{coverage} for additional details of the word vectors). Next, we create the vector of a subword as the average of the vector of the words that are connected with the subword:
\vspace{-2mm}
\begin{equation*}
    \vec{c} = \frac{1}{|N(c)|} \sum_{v \in N(c)} \boldsymbol{W}_{\{v\}}
\end{equation*}
\vspace{-5mm}

\noindent where $c$ is a subword in the graph and $N(c)$ 
is the set of
neighbors of $c$ in the graph
(these neighbors are $\in V$).
The intuition behind this calculation is that any words that include the same subword are related to the concept that the subword represents, and therefore those words should contribute to the representation of the subword.
If a subword in $V^s$ is not in the graph, we create its vector as zero. In this way, we create vectors for all subwords in $V^s$. We refer to the created subword vectors as $\boldsymbol{U}^s$. 

\paragraph{Step 3.} We create subword vectors for all subwords in $V^t$ in the same way as described in Step 2, using target decoder $\texttt{TOK}^t$, all words in $V$, and the multilingual word vectors $\boldsymbol{W}$.
The created subword vectors are denoted as $\boldsymbol{U}^t$. Note that $\boldsymbol{U}^t$ and $\boldsymbol{U}^s$ are in the same vector space as $\boldsymbol{W}$, because both of them are created based on $\boldsymbol{W}$.

\paragraph{Step 4.} We then leverage the source coordinates
$\boldsymbol{F}^s$, source-language subword vectors
$\boldsymbol{U}^s$ and target-language subword vectors
$\boldsymbol{U}^t$ to initialize target coordinates
$\boldsymbol{F}^{t}$. To begin with, we deal with the
subwords shared by  $V^s$ and $V^t$. For these subwords, we simply copy their coordinates from $\boldsymbol{F}^s$ to $\boldsymbol{F}^t$, which is also done by \citet{dobler2023focus}. For the remaining subwords, which are probably from new languages and not covered by $V^s$, we follow \wechsel \citep{minixhofer-etal-2022-wechsel} to find a good initialization based on similarity. Specifically, for each subword $x \in V^s$ and each subword $y \in V^t$, we calculate the cosine similarity between $x$ and $y$ in the subword vector space:
\vspace{-2mm}
\begin{equation*}
s_{(x, y)} = \text{cos-sim}({\boldsymbol{U}^s_{\{x\}}},\,\,{\boldsymbol{U}^t_{\{y\}}})
\end{equation*}
\vspace{-6mm}

\noindent The coordinate of each non-shared subword in $V^{t}$ is finally initialized as a convex combination of source-language coordinates in $\boldsymbol{F}^s$:
\vspace{-2mm}
\begin{equation*}
\boldsymbol{F}^{t}_{\{y\}} = \frac{\sum_{x \in \mathbb{N}(y)} \text{exp}(s_{(x, y)} / \tau)\cdot \boldsymbol{F}^{s}_{\{x\}}}{\sum_{x^\prime \in \mathbb{N}(y)} \text{exp}(s_{(x^\prime, y)} / \tau)}
\end{equation*}
\vspace{-5mm}

\noindent where $\mathbb{N}(y)$ is the set of $k$ nearest source-language subwords of the target-language subword $y$ and $\tau$ is the temperature (we set $k=10$ and $\tau=0.1$ by default, following \citet{minixhofer-etal-2022-wechsel} who report  the optimal choices in their experiments). In case the vector of a subword $y$ in $\boldsymbol{U}^t$ is zero, we randomly initialize its coordinate $\boldsymbol{F}^{t}_{\{y\}}$ from a Gaussian distribution $\mathcal{N}(\mathbb{E}[\boldsymbol{F}^{s}], \text{Var}[\boldsymbol{F}^{s}])$. Note that $\boldsymbol{F}^{t}$ is roughly in the embedding space of $\boldsymbol{F}^{s}$, instead of in the vector space of $\boldsymbol{U}^s$ and $\boldsymbol{U}^t$.

\paragraph{Step 5.} We finally assemble a target model by
using the transformer body of the source PLM (all parameters except for its subword embeddings), the primitive embeddings $\boldsymbol{P}$, and the initialized target coordinates $\boldsymbol{F}^{t}$. The dimension of $\boldsymbol{F}^{t}$ is the same as the transformer body if no matrix factorization is applied, otherwise, we need to up-project the coordinates with $\boldsymbol{P}$ to suit the hidden dimension of the transformer body. In this way, we transform a source PLM into a multilingual model that has fewer parameters, which serves as a good start for efficient multilingual continued pretraining.

\section{Experiments}
\subsection{Setups}
We use a SentencePiece \citep{kudo-richardson-2018-sentencepiece} tokenizer that has a vocabulary size of 401K as the target tokenizer. The vocabulary is merged from the subwords in XLM-R \citep{conneau-etal-2020-unsupervised} and new subwords learned from the Glot500-c corpus \citep{imani-etal-2023-glot500} (See \secref{corpus} for details of the Glot500-c corpus.). The target tokenizer is the same as the tokenizer used in Glot500-m \citep{imani-etal-2023-glot500}. We then created 8 models using \frameworkname framework as follows: 

\paragraph{\frameworkname-mono-xxx:} we construct target models by \frameworkname using English RoBERTA \citep{liu2019roberta} as the source model. xxx denotes the latent dimension used in the factorization, where singular value decomposition (SVD) is used and top-$k$ eigenvalues / eigenvectors are selected. We use four different dimensions: 100, 200, 400 and 768. When the dimension is 768, no matrix factorization is applied. The vocabulary and the tokenizer are the same as Glot500-m. Then we continued pretrain these assembled models on the Glot500-c corpus. 
% See \secref{corpus} for details of Glot500-c.
    
\paragraph{\frameworkname-multi-xxx:} we use the same setting as \frameworkname-mono-xxx to construct target models (latent dimension: 100, 200, 400, 768), where XLM-R is used as the source model. Then we continued pretrain these models on the Glot500-c corpus.\\

The model architecture of \frameworkname-mono-768 and \frameworkname-multi-768 is the same as Glot500-m, where the embeddings are tied with the parameters of the language modeling head. For lower-dimensional models, two matrices are used to map the representation back to vocabulary space for masked language modeling. The parameters of the two matrices are tied to the primitive embeddings and target coordinates. We continued pretrain all models using MLM objective and follow the training hyperparameters used by \citet{imani-etal-2023-glot500}. 
Each training step contains an effective batch of 384 samples randomly picked
from all language-scripts\footnote{A language-script is a combination of ISO 639-3 and script, which is used by the Glot500-c corpus.}. We refer to the languages that XLM-R covers as \textbf{head} languages and the remaining languages as \textbf{tail} languages. We store checkpoints for each model every 10K steps and apply early stopping with the best average performance on downstream tasks.
We train all models on \textbf{four} NVIDIA RTX A6000 GPUs for a maximum of four weeks.
% different from \textbf{eight} GPUs used by \citet{imani-etal-2023-glot500}. 
See \secref{hyperparam} for a detailed description of hyperparameter settings of continued pretraining and evaluation.

\begin{table}
    \setlength{\belowcaptionskip}{-0.4cm}
    \small
    \centering
    \def\tablesep{0.05cm}
    \begin{tabular}{
      @{\hspace{\tablesep}}l@{\hspace{\tablesep}}|
      @{\hspace{\tablesep}}r@{\hspace{\tablesep}}
      @{\hspace{\tablesep}}r@{\hspace{\tablesep}}
      @{\hspace{\tablesep}}r@{\hspace{\tablesep}}
      @{\hspace{\tablesep}}r@{\hspace{\tablesep}}
    }
    & $D^\prime$=100 & $D^\prime$=200 & $D^\prime$=400 & $D$=768\\
    \midrule
    Model Params.  & 126M  & 167M & 247M & 395M \\
    Embedding Params.  & 40M  & 80M & 161M & 309M \\
    \end{tabular}
    \caption{Model parameters under different latent dimensions. When $D^\prime$=100, 200, or 400, each corresponds to two \frameworkname-initialized models (based on RoBERTa or XLM-R). $D$=768 not only corresponds to \frameworkname-768, but also baselines RoBERTa-rand and XLM-R-rand, as they have the same architecture. By decreasing latent dimensions, the model parameters decrease drastically.}
    \label{tab:model_size}
\end{table}

\subsection{Baselines}
We consider the following baselines for comparison with \frameworkname (see Table \ref{tab:model_size} for the number of parameters under different latent embedding dimensions):
\paragraph{RoBERTa} A monolingual PLM trained on English corpus \citep{liu2019roberta}. Its embeddings and tokenizer do not cover most of the new subwords of our models. The vocabulary size is 50K. 

\paragraph{RoBERTa-rand} We replace the embeddings of RoBERTa with new embeddings (the vocabulary size is 401K, the same as \frameworkname-mono-768), which are constructed by copying the shared subwords and \textbf{randomly} initializing the embeddings of remaining subwords not covered by RoBERTa from a Gaussian distribution with a mean and variance of the original RoBERTa embeddings, similar to \citet{minixhofer-etal-2022-wechsel}.  Glot500-m tokenizer is used for tokenization. We then continued pretrain it on Glot500-c with the same hyperparameters.

\paragraph{XLM-R} A strong multilingual PLM trained on 100 languages \citep{conneau-etal-2020-unsupervised}. We use the \textbf{base} version, where the embedding dimension is 768. The vocabulary size is 250K.

\paragraph{XLM-R-rand} Similar to RoBERTa-rand, this model extends the vocabulary from XLM-R, and the embeddings of subwords not covered by XLM-R are randomly initialized from a Gaussian distribution with a mean and variance of the original XLM-R embeddings.\footnote{The model is named Glot500-m in \citet{imani-etal-2023-glot500}. To be consistent with other names used in this paper, we call it XLM-R-rand. All models are trained on the same infrastructure for a strictly controlled experimental setting.}  Glot500-m tokenizer is used for tokenization. The model is then continued pretrained on Glot500-c with the same hyperparameters.

\begin{table*}[ht]
    \setlength{\belowcaptionskip}{-0.4cm}
    \footnotesize
    \centering
    \setlength{\tabcolsep}{1.7mm}{}
    \begin{tabular}{lrrrrrrrrrrrrrrr}
        \toprule
        & \multicolumn{3}{c}{SR-B} & \multicolumn{3}{c}{SR-T} & \multicolumn{3}{c}{Taxi1500} & \multicolumn{3}{c}{NER} & \multicolumn{3}{c}{POS}\\
        \cmidrule(lr){2-4} \cmidrule(lr){5-7} \cmidrule(lr){8-10} \cmidrule(lr){11-13} \cmidrule(lr){14-16}
        & tail & head & all & tail & head & all & tail & head & all & tail & head & all & tail & head & all\\
        % \midrule
        % XLM-V & 10.1 & \underline{60.3} & 22.9 & 39.3 & \textbf{78.2} & \textbf{67.1} & 17.7 & 61.1 & 28.7 & 53.2 & 65.3 & 59.8 & 45.4 & \textbf{76.6} & 67.0 \\
        \midrule
        RoBERTa & 3.2 & 3.9 & 3.4 & 8.1 & 4.9 & 5.8 & 5.5 & 6.9 & 5.8 & 30.4 & 26.4 & 28.2 & 21.1 & 28.6 & 26.3 \\
        RoBERTa-rand & 11.0 & 14.7 & 11.9 & 24.9 & 20.9 & 22.0 & 14.2 & 19.1 & 15.5 & 52.1 & 49.8 & 50.8 & 47.1 & 61.4 & 57.0 \\
        \frameworkname-mono-100 & 13.1 & 20.3 & 14.9 & 26.8 & 26.5 & 26.6 & 15.8 & 24.8 & 18.1 & 53.3 & 52.6 & 52.9 & 50.6 & 64.8 & 60.4 \\
        \frameworkname-mono-200 & \underline{16.1} & \underline{25.9} & \underline{18.6} & \underline{33.2} & \underline{34.3} & \underline{33.9} & \underline{29.8} & \underline{37.0} & \underline{31.6} & \underline{55.8} & \underline{56.1} & \underline{56.0} & 49.0 & 66.1 & 60.8 \\
        \frameworkname-mono-400 & \textbf{25.4} & \textbf{40.4} & \textbf{29.2} & \textbf{41.6} & \textbf{48.7} & \textbf{46.7} & \textbf{35.1} & \textbf{46.4} & \textbf{37.9} & \textbf{58.2} & \textbf{59.0} & \textbf{58.6} & \textbf{57.0} & \textbf{70.6} & \textbf{66.4} \\
        \frameworkname-mono-768 & 16.0 & 23.6 & 17.9 & 28.6 & 28.5 & 28.6 & 22.1 & 28.9 & 23.8 & 54.8 & 55.3 & 55.1 & \underline{51.7} & \underline{66.7} & \underline{62.1} \\
        \midrule
        XLM-R & 7.4 & 54.2 & 19.3 & 32.6 & 66.2 & 56.6 & 15.5 & 59.8 & 26.7 & 47.6 & 61.8 & 55.3 & 42.1 & \textbf{76.1} & 65.6 \\
        XLM-R-rand & 38.6 & \underline{60.4} & 44.2 & \underline{55.6} & \underline{69.7} & \underline{65.7} & 47.0 & \underline{59.9} & 50.3 & 60.3 & 62.3 & 61.4 & 60.6 & 74.9 & 70.5 \\
        \frameworkname-multi-100 & 33.0 & 49.7 & 37.3 & 54.9 & 63.8 & 61.3 & 50.5 & 56.7 & 52.1 & 58.6 & 59.8 & 59.2 & 60.4 & 73.9 & 69.7 \\
        \frameworkname-multi-200 & 39.4 & 57.0 & 43.9 & 51.8 & 61.1 & 58.5 & 49.0 & 54.9 & 50.5 & 59.5 & 61.4 & 60.6 & 60.5 & 74.9 & 70.5  \\
        \frameworkname-multi-400 & \textbf{44.5} & 60.0 & \underline{48.5} & 54.8 & 64.7 & 61.8 & \underline{51.9} & 59.3 & \underline{53.8} & \textbf{62.5} & \textbf{64.0} & \textbf{63.3} & \textbf{63.2} & 75.4 & \underline{71.6} \\
        \frameworkname-multi-768 & \underline{43.8} & \textbf{62.7} & \textbf{48.7} & \textbf{56.1} & \textbf{70.4} & \textbf{66.3} & \textbf{54.3} & \textbf{63.8} & \textbf{56.7} & \underline{60.6} & \underline{63.9} & \underline{62.4} & \underline{62.4} & \underline{75.8} & \textbf{71.7} \\
        \bottomrule
    \end{tabular}
    \caption{Performance of the models initialized with \frameworkname and baselines on five multilingual tasks across 5 seeds. We report the performance as an average over head, tail, and all language-scripts for each model. Models initialized with \frameworkname constantly perform better than baselines.
    \textbf{Bold} (\underline{underlined}): best (second-best) result per controlled group.}
    \label{tab:results_with_cp}
\end{table*}

\subsection{Downstream Tasks}
\paragraph{Sentence Retrieval.} We consider two datasets: Tatoeba \citep{artetxe-schwenk-2019-massively} (SR-T) and Bible (SR-B). We select up to 1,000 English-aligned sentences for SR-T, following the same setting used by \citet{hu2020xtreme}. For SR-B, we select up to 500 English-aligned sentences. We report the top-10 accuracy by finding the nearest neighbors of the representation of each English sentence. Following \citet{jalili-sabet-etal-2020-simalign}, the representations are calculated by taking the average of the contextualized word embeddings at the 8th layer. 

\paragraph{Sequence Labeling.} We consider two types of tasks: named entity recognition (NER) and Part-Of-Speech (POS) tagging. We use WikiANN dataset \citep{pan-etal-2017-cross} for NER and Universal Dependencies \citep{de-marneffe-etal-2021-universal} of version v2.11 for POS. We finetune the models only on the English train set, select the best model on the English dev set, and then report the zero-shot performance on the test sets of other languages. F1 scores are reported for both NER and POS.

\paragraph{Text Classification.} We use Taxi1500 \citep{ma2023taxi1500}, a text classification dataset that provides train/dev/test sets with 6 classes in more than 1,500 languages. Following \citet{imani-etal-2023-glot500}, we select a subset of languages (351) supported by the models for evaluation. Same as in NER and POS, we report the zero-shot performance (in F1 scores) using English as the source. 

\begin{figure*}[htbp]
\setlength{\belowcaptionskip}{-0.5cm}
\centering
\subfigure[Training loss]{
\begin{minipage}[t]{0.32\linewidth}
\centering
\includegraphics[width=2.05in]{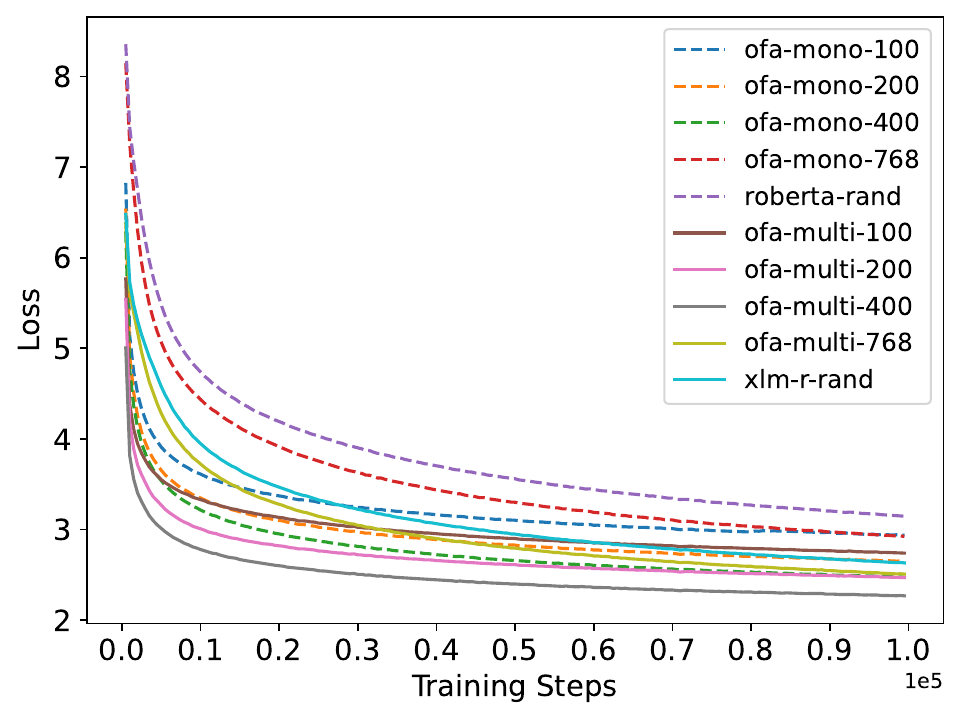}
\end{minipage}%
}%
\subfigure[SR-B]{
\begin{minipage}[t]{0.32\linewidth}
\centering
\includegraphics[width=2.05in]{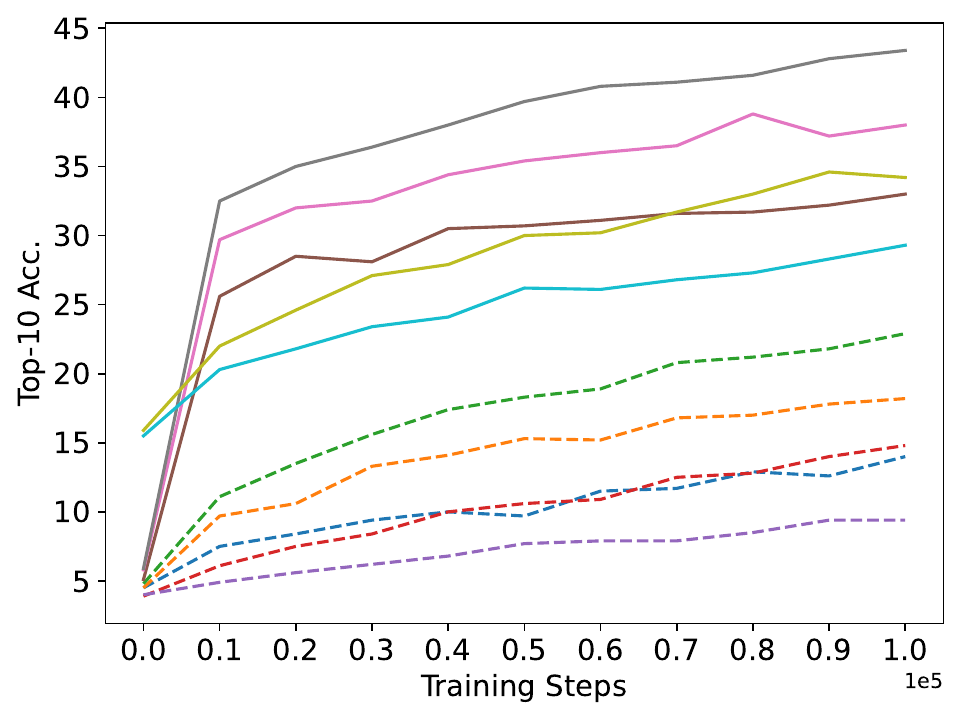}
\end{minipage}%
}%
\subfigure[SR-T]{
\begin{minipage}[t]{0.32\linewidth}
\centering
\includegraphics[width=2.05in]{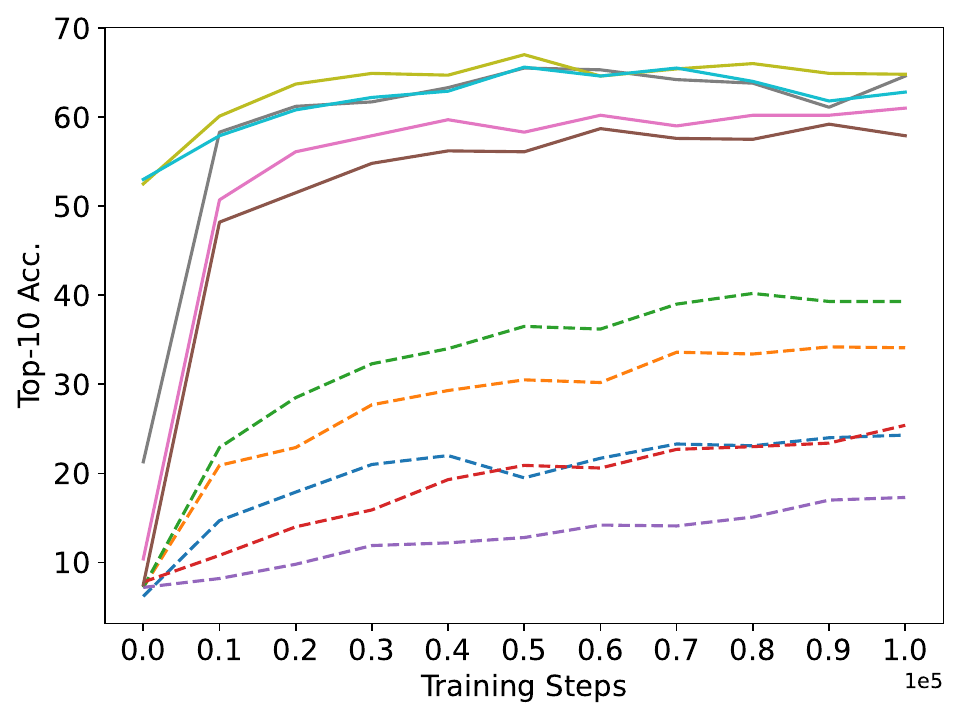}
\end{minipage}%
}
\\
\subfigure[Taxi1500]{
\begin{minipage}[t]{0.32\linewidth}
\centering
\includegraphics[width=2.05in]{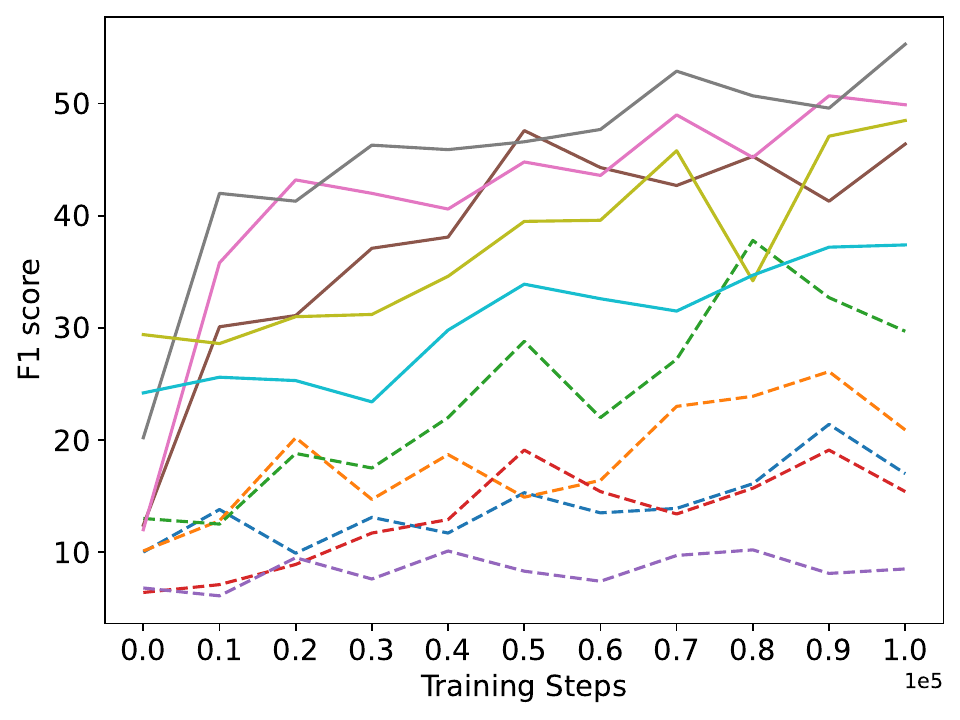}
\end{minipage}%
}%
\subfigure[NER]{
\begin{minipage}[t]{0.32\linewidth}
\centering
\includegraphics[width=2.05in]{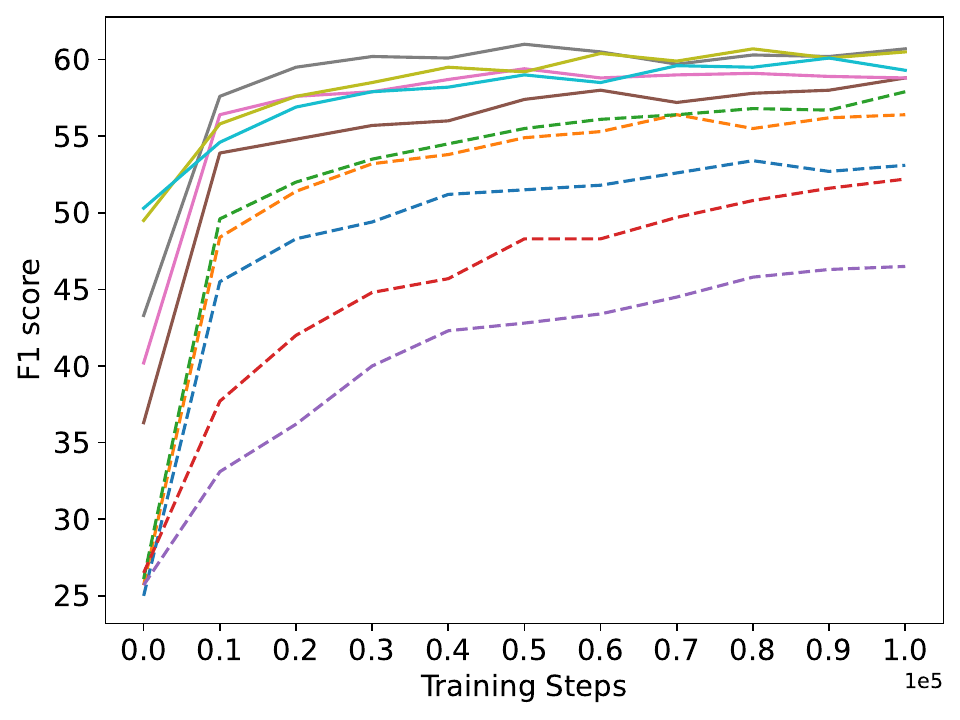}
\end{minipage}%
}%
\subfigure[POS]{
\begin{minipage}[t]{0.32\linewidth}
\centering
\includegraphics[width=2.05in]{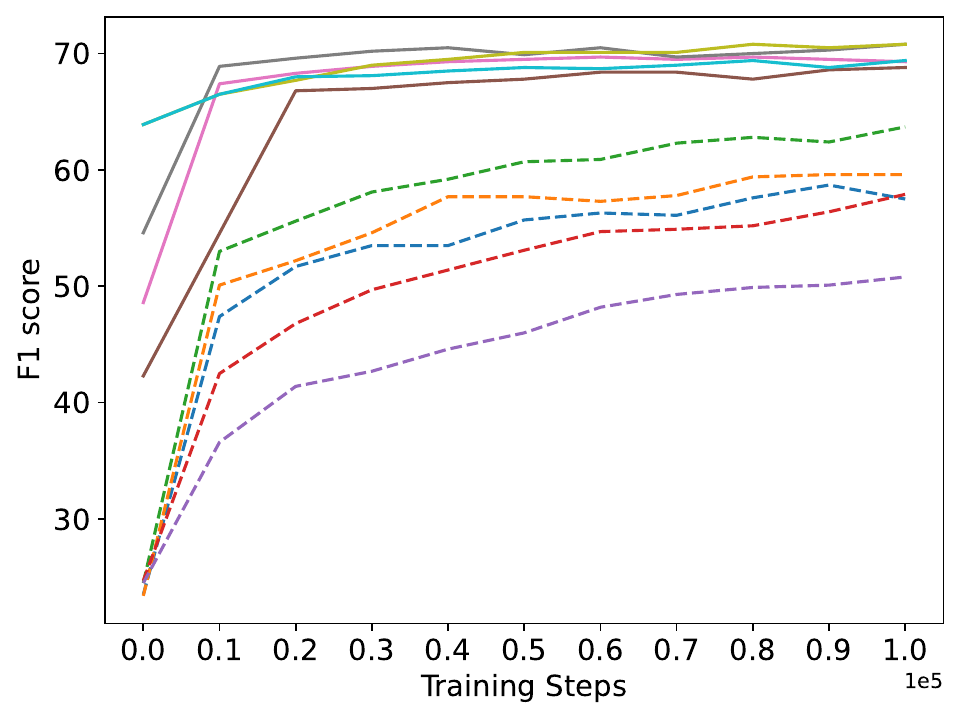}
\end{minipage}%
}%
\centering
\caption{The training loss as well as the performance on five downstream tasks from step 0 (without continued pretraining) to step 100K (10th checkpoints). We see that models initialized by \frameworkname converge faster than baseline models (RoBERTa-rand and XLM-R-rand) whose new subwords are randomly initialized during continued pretraining. For most of the downstream tasks, models with lower embedding dimensions can achieve better performance after only 10K steps compared with their full-dimensional counterparts (\frameworkname-mono-768 and \frameworkname-multi-768).}
\label{fig:outputs}
\end{figure*}

\begin{table}[h]
  \centering
  \small
  \setlength{\belowcaptionskip}{-0.5cm}
  \setlength{\tabcolsep}{1.2mm}{}
  \begin{tabular}{l|r|rr|rr}
\toprule
 & |L| & \#rand & \#\frameworkname-mono & \#rand & \#\frameworkname-multi\\
\midrule
SR-B & 369 & 0 & \textbf{369} & 23 & \textbf{346} \\
SR-T & 98 & 1 & \textbf{97} & 24 & \textbf{74} \\
Taxi1500 & 351 & 5 & \textbf{346} & 31 & \textbf{320} \\
NER & 164 & 10 & \textbf{154} & 27 & \textbf{137} \\
POS & 91 & 4 & \textbf{87} & 12 & \textbf{79} \\
\bottomrule
  \end{tabular}
  \caption{Number of languages in each downstream task that benefits from \frameworkname framework compared with randomly initializing the new subwords. |L| is the total number of languages for each task. \#rand (resp. \#\frameworkname): number of languages on which better performance is achieved by random initialization (resp. one of the latent dimensions in \frameworkname initialization) when using a monolingual (mono) or multilingual (multi) as the source PLM.
  \label{tab:whosbetter}}
\end{table}

\subsection{Results and Discussions}
Table \ref{tab:results_with_cp} shows the performance of the models initialized with \frameworkname and baselines with random initialization of new subword embeddings on five downstream tasks (see complete results for each language-script in \secref{complete_results}). Models initialized with \frameworkname demonstrate a consistent improvement on either head or tail languages compared with the baselines. Combined with Table \ref{tab:whosbetter}, we see that more languages benefit from \frameworkname initialization for both using the monolingual and multilingual PLM as the source model, which indicates an overall superiority of the \frameworkname initialization. 

When the source model is monolingual, with random initialization of unseen subwords, RoBERTa-rand just obtains 11.9, 22.0, and 15.5 on SR-B, SR-T, and Taxi1500 respectively (averaged overall), which are 6.0, 6.6, 8.3 lower than its counterpart \frameworkname-mono-768. In the sequence labeling task we also see similar improvement: \frameworkname-mono-768 achieves 4.3 and 5.1 better than RoBERTa-rand on NER and POS respectively. Such an increase is even higher when compared with RoBERTa, as RoBERTa is a monolingual model. When the source model is multilingual, models initialized with \frameworkname also achieve remarkable performance. \frameworkname-multi-768 achieves better performance than XLM-R on every task. Compared with XLM-R-rand, it also achieves better performance, which indicates the effectiveness of the initialization with the help of external multilingual embeddings.

The embedding dimension also plays a crucial role in the performance. Typically, we see an improvement in performance as we increase the latent dimension, particularly from 100 to 400 for both \frameworkname-mono and \frameworkname-multi models. This is expected as a larger dimension often induces better expressiveness. Nevertheless, the improvement from dimension 400 to 768, is not consistently large, and in some cases, it even leads to performance declines. For example, \frameworkname-mono-400 outperforms \frameworkname-mono-768 on all downstream tasks. We assume this is because a monolingual model with many parameters might not be easy to adapt to diverse languages. A smaller embedding dimension can ease the burden and facilitate the pretraining, thus achieving better performance. Similarly, \frameworkname-multi-400 is very competitive to \frameworkname-multi-768 (\frameworkname-multi-400 is even better on NER and POS). We attribute this to the ``redundancy'' of the embeddings in multilingual PLMs (see \secref{redundancy} for an analysis). By using factorization, we keep the most important information that is shared across languages. Thus there is a trade-off. When the dimension is very small, e.g., 100, there is a risk of information loss. However, with a moderate size, e.g., 400, the model is less redundant and equipped with enough expressiveness to achieve good performance.  

\section{Analysis}

\subsection{Continued training Progression}\seclabel{progression}

To analyze how different embedding dimensions and initialization methods can influence the continued training, we visualize the training loss of models that are initialized with \frameworkname and two baseline models, i.e., RoBERTa-rand and XLM-R-rand. In addition, we evaluate all these models on five downstream tasks at 10K-step intervals until 100K steps. The results are shown in Figure \ref{fig:outputs}. From Fig. \ref{fig:outputs} (a), when the embedding dimension is 768, the models initialized with \frameworkname converge faster compared with the models being randomly initialized, regardless of whether the source model is monolingual or multilingual. The faster convergence is also related to the performance, as \frameworkname-mono-768 (resp.\  \frameworkname-multi-768) constantly performs better than RoBERTa-rand (resp.\  XLM-R-rand) throughout steps for all tasks. This indicates that \frameworkname, which explicitly leverages information encoded in source PLM embeddings and external multilingual word vectors, is superior to random initialization.  

\begin{table}
\setlength{\belowcaptionskip}{-0.4cm}
\centering
\small
\setlength\tabcolsep{5pt}
\begin{tabular}{lcrr}
        \toprule
        Models & best-checkpoint &  avg. $T$ & C.F.\\
        \midrule
        \multirow{1}{*}{\frameworkname-mono-100} 
        & 110K & 3.8h & 21.7 \\
        \multirow{1}{*}{\frameworkname-mono-200} 
        & 120K & 3.9h & 24.3 \\
        \multirow{1}{*}{\frameworkname-mono-400} 
        & 230K & 4.3h & 51.3 \\
        \multirow{1}{*}{\frameworkname-mono-768} 
        & 250K & 4.7h & 60.9 \\
        \multirow{1}{*}{RoBERTa-rand} 
        & 270K & 4.7h & 65.8 \\
        \midrule
        \multirow{1}{*}{\frameworkname-multi-100} 
        & 290K & 3.8h & 57.1 \\
        \multirow{1}{*}{\frameworkname-multi-200} 
        & 280K & 3.9h & 56.6 \\
        \multirow{1}{*}{\frameworkname-multi-400} 
        & 260K & 4.3h & 58.0 \\
        \multirow{1}{*}{\frameworkname-multi-768} 
        & 450K & 4.7h & 110.0 \\
        \multirow{1}{*}{XLM-R-rand} 
        & 560K & 4.7h & 136.4 \\
        \bottomrule
    \end{tabular}
\caption{\label{tab:additional_info}
Additional information: best checkpoint, average training time (avg. $T$) spent per 10K steps until the best checkpoint, and carbon footprint (C.F.: in kg of $\text{CO}_2$ eq.) of different models in continued pretraining.
}
\end{table}

We also observe models with smaller dimensions tend to learn information faster in the initial steps, indicated by the speed of MLM loss drop. As explained earlier, smaller dimensions mean fewer parameters which eases the burden in continued pretraining, especially when the source model is monolingual. On the other hand, faster learning speed explains why models with smaller dimensions generally perform better than their full-dimensional counterparts (\frameworkname-mono-768 or \frameworkname-multi-768) in the early training phase. For example, with only 167M parameters, \frameworkname-multi-200 achieves better or very close performance on each task compared with \frameworkname-multi-768, which is two times larger. We also observe that all models, especially \frameworkname-multi models, quickly reach a performance plateau on NER and POS tasks. This aligns with the finding that syntactic knowledge is acquired rapidly in the training progression \citep{blevins-etal-2022-analyzing,mullereberstein2023subspace}. This also suggests that sequence labeling might be a straightforward task where the model can transfer prevalent classes such as \textit{verb} and \textit{noun}, possibly through shared vocabulary \citep{imani-etal-2023-glot500}.

Combined with the analysis above, better initialization and smaller embedding dimensions enable an efficient multilingual continued pretraining and better performance in downstream tasks with fewer training steps. Lightweight models also reduce GPU consumption and allow for larger batch sizes. Therefore, the proposed \frameworkname framework can be very useful where a limited computation budget is presented, e.g., in most laboratories or institutions. 

In addition, as there are recent concerns regarding the environmental impact of training or operating LMs \citep{bender2021dangers,rae2021scaling,weidinger2022taxonomy}, we also report some related statistics when continued pretraining our models in Table \ref{tab:additional_info}. There are two benefits of using \frameworkname with factorized embedding parameterization:  (1) the average training time per 10K steps is shortened and (2) overall less training time is required to reach the best checkpoints compared to the random baseline. Considering that there is no huge difference in terms of the performance in downstream tasks, initializing by \frameworkname with lower embedding dimensions can largely reduce the carbon emissions\footnote{Estimations were conducted using the \href{https://mlco2.github.io/impact\#compute}{MachineLearning Impact calculator} presented in \citep{lacoste2019quantifying}.} and therefore is more environmentally friendly.

\begin{table}
\setlength{\belowcaptionskip}{-0.4cm}
\centering
\small
\setlength\tabcolsep{1.5pt}
\begin{tabular}{lcrrrrr}
        \toprule
        Models & Settings & SR-B & SR-T & Taxi1500 & NER & POS\\
        \midrule
        \multirow{2}{*}{\frameworkname-mono-100} 
        & w/o & 4.5 & 6.2 & 10.0 & 25.0 & 23.5 \\
        & w/ & \textbf{14.9} & \textbf{26.6} & \textbf{18.1} & \textbf{52.9} & \textbf{60.4} \\ 
        % \midrule
        \multirow{2}{*}{\frameworkname-mono-200} 
        & w/o & 4.5 & 7.2 & 10.1 & 25.7 & 23.4 \\
        & w/ & \textbf{18.6} & \textbf{33.9} & \textbf{31.6} & \textbf{56.0} & \textbf{60.8} \\ 
        % \midrule
        \multirow{2}{*}{\frameworkname-mono-400} 
        & w/o & 4.8 & 7.2 & 13.0 & 26.1 & 24.5 \\
        & w/ & \textbf{29.2} & \textbf{46.7} & \textbf{37.9} & \textbf{58.6} & \textbf{66.4} \\ 
        % \midrule
        \multirow{2}{*}{\frameworkname-mono-768} 
        & w/o & 3.9 & 7.8 & 8.2 & 26.5 & 24.7 \\
        & w/ & \textbf{17.9} & \textbf{28.6} & \textbf{23.8} & \textbf{55.1} & \textbf{62.1} \\ 
        % \midrule
        \midrule
        \multirow{2}{*}{\frameworkname-multi-100} 
        & w/o & 5.1 & 7.5 & 12.4 & 36.3 & 42.3 \\
        & w/ & \textbf{37.3} & \textbf{61.3} & \textbf{52.1} & \textbf{59.2} & \textbf{69.7} \\ 
        % \midrule
        \multirow{2}{*}{\frameworkname-multi-200} 
        & w/o & 5.7 & 10.4 & 12.0 & 40.2 & 48.6 \\
        & w/ & \textbf{43.9} & \textbf{58.5} & \textbf{50.5} & \textbf{60.6} & \textbf{70.5} \\ 
        % \midrule
        \multirow{2}{*}{\frameworkname-multi-400} 
        & w/o & 5.9 & 21.3 & 20.2 & 43.3 & 54.6 \\
        & w/ & \textbf{48.5} & \textbf{61.8} & \textbf{53.8} & \textbf{63.3} & \textbf{71.6} \\ 
        % \midrule
        \multirow{2}{*}{\frameworkname-multi-768} 
        & w/o & 15.9 & 52.5 & 29.4 & 49.5 & 63.9 \\
        & w/ & \textbf{48.7} & \textbf{66.3} & \textbf{56.7} & \textbf{62.4} & \textbf{71.7} \\ 
        \bottomrule
    \end{tabular}
\caption{\label{tab:continue_pretraining}
Performance of models initialized with \frameworkname under settings of w/o and w/ continued pretraining. Continued pretraining largely improves the performance.
}
\end{table}

\subsection{Influence of Continued Pretraining}

Continued pretraining has a different impact on models with different latent embedding dimensions for different downstream tasks. Therefore, we compare how the model performance varies with or without continued pretraining, as shown in Table \ref{tab:continue_pretraining}.

\begin{table*}[ht]
    \scriptsize
    \centering
    \setlength{\belowcaptionskip}{-0.2cm}
    \setlength{\tabcolsep}{1.0mm}{}
    \begin{tabular}{lrrrrrrrrr}
\toprule
 & (indo1319, 93) & (atla1278, 69) & (aust1307, 55) & (turk1311, 23) & (sino1245, 23) & (maya1287, 15) & (afro1255, 12) & (other, 79) & (all, 369) \\
\midrule
RoBERTa & 4.8 & 3.0 & 3.3 & 2.3 & 3.0 & 2.5 & 2.7 & 2.8 & 3.4 \\
RoBERTa-rand & 17.8 & 10.0 & 14.7 & 10.1 & 8.8 & 7.3 & 7.0 & 7.8 & 11.9 \\
\frameworkname-mono-100 & 22.6 & 13.0 & 16.9 & 13.3 & 9.8 & 7.4 & 8.3 & 10.6 & 14.9 \\
\frameworkname-mono-200 & \underline{28.7} & \underline{15.8} & 20.1 & \underline{19.5} & 13.3 & \underline{8.2} & 10.8 & \underline{12.5} & \underline{18.6} \\
\frameworkname-mono-400 & \textbf{44.1} & \textbf{25.3} & \textbf{30.5} & \textbf{34.0} & \textbf{21.4} & \textbf{10.9} & \textbf{17.4} & \textbf{20.4} & \textbf{29.2} \\
\frameworkname-mono-768 & 26.3 & 15.6 & \underline{20.8} & 18.7 & \underline{14.3} & 7.9 & \underline{11.0} & 11.9 & 17.9 \\
\midrule
XLM-R & 41.9 & 5.5 & 14.5 & 22.3 & 9.0 & 3.8 & 13.0 & 14.1 & 19.3 \\
XLM-R-rand & 61.3 & 38.9 & 44.9 & 62.2 & 33.9 & 15.0 & 33.1 & 33.1 & 44.2 \\
\frameworkname-multi-100 & 53.4 & 35.8 & 36.9 & 52.5 & 27.2 & 11.3 & 24.2 & 25.2 & 37.3 \\
\frameworkname-multi-200 & 60.3 & 41.8 & 43.3 & 61.4 & 34.3 & 15.1 & 31.5 & 31.9 & 43.9 \\
\frameworkname-multi-400 & \underline{63.9} & \textbf{46.7} & \underline{48.0} & \underline{65.9} & \underline{39.4} & \textbf{19.6} & \textbf{36.0} & \underline{37.2} & \underline{48.5} \\
\frameworkname-multi-768 & \textbf{64.6} & \underline{46.5} & \textbf{48.3} & \textbf{66.7} & \textbf{39.5} & \underline{17.7} & \underline{35.4} & \textbf{37.4} & \textbf{48.7} \\
        \bottomrule
    \end{tabular}
    \caption{Aggregated performance of the models for 7 major language families on \textbf{SR-B}. We report the average performance for \textbf{indo1319} (Indo-European), \textbf{atla1278 } (Atlantic-Congo), \textbf{aust1307} (Austronesian), \textbf{turk1311} (Turkic), \textbf{sino1245} (Sino-Tibetan), \textbf{maya1287} (Mayan), and \textbf{afro1255} (Afro-Asiatic). We classify the remaining languages into the group ``\textbf{other}''. In addition, we report the average over all languages (group ``\textbf{all}'').  The number of languages in that family is shown in the parentheses.
    \textbf{Bold} (\underline{underlined}): best (second-best) result for each task.}
    \label{tab:family_sr_b}
\end{table*}

Although most models without continued pretraining perform generally badly, we see some exceptions. For example, \frameworkname-multi-768 achieves more than 52.5 accuracy in SR-T, while only 15.9 in SR-B. The major reason is that SR-B contains many tail language-scripts that are not covered by XLM-R. On the contrary, SR-T contains many head languages
and many of the other languages are similar to those head languages. 
We also notice that the continued pretraining has less impact on sequence labeling tasks, i.e., NER and POS, where the model can use the knowledge already encoded in its parameters to perform well in English, and then transfer to other languages through shared vocabulary, or the already existing crosslinguality when the source model is multilingual. 

When the source model is monolingual, the performance without continued pretraining is bad no matter which embedding dimension is used. However, the higher-dimension model achieves consistently better performance than lower-dimension ones when the source model is multilingual. This can be explained by the fact that the source multilingual model already has strong crosslinguality and a higher dimension can better restore the original information encoded in XLM-R's embedding matrix. Nevertheless, the benefits of higher dimensions diminish after continued pretraining. Combined with Figure \ref{fig:outputs}, we see that even the smallest model, i.e., \frameworkname-multi-100, quickly surpasses \frameworkname-multi-768 in SR-B and Taxi500 tasks after 10K training steps. We therefore could conclude that the models initialized with \frameworkname could quickly adapt to new languages in the continued pretraining, especially when the source model is already multilingual.

\subsection{Performance across Language Families}
The aggregate results shown in Table \ref{tab:results_with_cp} reflect that \frameworkname can improve the overall performance. However, the results can potentially hide some information such as for what kind of language families and / or scripts \frameworkname works better or worse. Thus we also report the aggregated performance for major language families in \textbf{SR-B} that covers the most languages among our downstream tasks. The results are shown in Table \ref{tab:family_sr_b} (see aggregated results for different scripts and other tasks in \secref{complete_results_family_script}).

It can be seen that all variants with \frameworkname initialization consistently outperform the random initialization baselines across all language families when using RoBERTa as the source model. Similarly, when the latent dimension is larger or equal to 400, models with \frameworkname initialization beat the counterparts across all language families. These findings indicate \frameworkname's superiority is not limited to certain language families. In addition, we find the performance difference between \frameworkname-multi-400 and \frameworkname-multi-768 is small across language families, which further indicates that reducing the dimension of embeddings is effective in continued pretraining.

\section{Conclusion}
In this work, we present \frameworkname, a framework that wisely initializes unseen subword embeddings with factorized embedding parameterization for efficient large-scale multilingual continued pretraining.
We conduct extensive and strictly controlled experiments by continued pretraining models that are initialized from monolingual or multilingual PLMs. We evaluate these models on a wide range of downstream tasks. We show that models initialized with \frameworkname enjoy faster convergence during training and achieve competitive or better performance on downstream tasks, compared with the baselines where embeddings of new subwords are randomly initialized. We also show that with smaller embedding dimensions, the continued pretraining is further facilitated: training time is shortened and models achieve better performance in the early training phase. Therefore, this work contributes to efficient large-scale multilingual continued pretraining.
% and \frameworkname is especially helpful to laboratories or institutions with a limited computation budget.

\section*{Limitations}

% not applied to other types of models such as decoder-only or encoder-decoder model
In this work, we apply \frameworkname to two models, RoBERTa, a monolingual PLM, and XLM-R, a multilingual PLM, and show the superiority of the proposed initialization method compared to the random initialization. However, both are encoder-only models and they are pretrained / continued pretrained only using the MLM objective. Theoretically, this approach should be able to extend to other types of models, e.g., decoder-only and encoder-decoder models, or other types of training objectives, e.g., next-word prediction or translation objectives, since our approach is \textbf{only related to the initialization stage} of continued pretraining and not restricted to any model architectures or training objectives. We do not try all possibilities in terms of architectures / objectives as that is not the major focus of this work, and we have a limited computation budget. We would leave such exploration using \frameworkname in different architectures / objectives for future research in the community.

% catastrophic forgetting the initial knowledge, maybe more active forgetting into the the training.
Another possible limitation is that, while we inject external knowledge into the subword embeddings before continued pretraining, such knowledge may diminish due to catastrophic forgetting \citep{Kirkpatrick2017catastrophic}. That is, due to continued pretraining, the model gradually loses the initial knowledge. This is not wanted and we would expect methods such as active forgetting \citep{chen2023improving} could alleviate the problem by restoring the constructed embeddings from \frameworkname every certain step in the continued pretraining. However, this again is not the major focus of this paper and we would call for exploration in this direction. 

% \section*{Ethics Statement}
% % Scientific work published at EMNLP 2023 must comply with the \href{https://www.aclweb.org/portal/content/acl-code-ethics}{ACL Ethics Policy}. We encourage all authors to include an explicit ethics statement on the broader impact of the work, or other ethical considerations after the conclusion but before the references. The ethics statement will not count toward the page limit (8 pages for long, 4 pages for short papers).

\section*{Acknowledgement}
We appreciate Fabian David Schmidt's suggestion to filter the unnecessary tokens in MLM modeling and re-estimate the training time of the models. We would also like to thank the reviewers for their constructive feedback. This work was funded by the European Research Council (grant \#740516).

\bibliography{anthology,custom}
\bibliographystyle{acl_natbib}

\appendix

%TODO maybe add a section to introduce the multilingual embeddings

\section{Glot500-c}\seclabel{corpus}
The Glot500-c corpus \citep{imani-etal-2023-glot500}\footnote{\url{https://github.com/cisnlp/Glot500}} contains 511 languages in 30 different scripts. The total number of sentences is 1.5B and the median number of sentences per language-script is 120K. Because some languages can be written in multiple scripts, the corpus treats each \textbf{language-script} as a separate entity. For example, Tajik-Cyrillic and Tajik-Arabic will be considered as different entities as there are two different scripts used for Tajik in the corpus. The corpus is divided into train/dev/test sets for each language. Dev and test sets have 1000 sentences. Same as \citep{imani-etal-2023-glot500}, we only use the training data to continued pretrain all of our models.

\section{Detailed Hyperparameters}\seclabel{hyperparam}
\subsection{Continued Pretraining}
We continued pretrain both the baseline models (RoBERTa-rand and XLM-R-rand) and models initialized with \frameworkname using basically the same hyperparameters as used in \citet{imani-etal-2023-glot500}. Specifically, we use MLM objective with the standard mask rate of 15\%. We use Adam optimizer \citep{ba2015adam} with $(\beta_1, \beta_2) = (0.9, 0.999)$ and $\epsilon = \text{1e-6}$. The initial learning rate is set to 5e-5. The effective batch size is set to 384. Each batch contains training samples concatenated up to the maximum sequence length of 512 and randomly picked from all language-scripts in the Glot500-c corpus. The only difference from ours to \citet{imani-etal-2023-glot500} is that we use \textbf{four} RTX A6000 GPUs while they use \textbf{eight} RTX A6000 GPUs. Therefore, we set the per-GPU batch to 12, and the gradient accumulation to 8, fulfilling $4 \times 12 \times 8 = 384$. The gradient accumulation in \citet{imani-etal-2023-glot500} is set to 4, as they use four more GPUs. We use FP16 training (mixed precision \citep{paulius2018mixed}). The different gradient accumulation and usage of mixed-precision might be the reason why the performance of our baseline XLM-R-rand is slightly different from the performance reported in \citet{imani-etal-2023-glot500}.
The continue-pretraining is done using scripts adapted from HuggingFace\footnote{\url{https://huggingface.co/}}.

\subsection{Downstream Tasks}

\begin{table}[t]
  \small
	\centering
	\def\tablesep{0.2cm}
\begin{tabular}{
  @{\hspace{\tablesep}}l@{\hspace{\tablesep}}|
  @{\hspace{\tablesep}}r@{\hspace{\tablesep}}
  @{\hspace{\tablesep}}r@{\hspace{\tablesep}}
  @{\hspace{\tablesep}}r@{\hspace{\tablesep}}
  @{\hspace{\tablesep}}c@{\hspace{\tablesep}}
}
 & |head| & |tail| & \#class & measure (\%) \\
  \midrule
SR-B    & 94 & 275 & - & top-10 Acc. \\
SR-T    & 70 & 28  & - & top-10 Acc. \\
Taxi1500    & 89 & 262 & 6 & F1 score \\
NER & 89 & 75 & 7 & F1 score \\
POS & 63 & 28 & 18 & F1 score \\
  \end{tabular}
  \caption{Downstream tasks and measures. |head| (resp.\  |tail|): head (resp.\  tail) language-scripts according to \citet{imani-etal-2023-glot500} (a language-script is head if it is covered by XLM-R, otherwise it is tail); \#class: the number of the categories if it is a (sequence-level or token-level) classification task.}
  \label{evaluationmetrics}
\end{table}

The outline of the evaluation is shown in Table \ref{evaluationmetrics}. We introduce the detailed hyperparameters used for each downstream task in the following.

\paragraph{SR-B.} We use up to 500 English-aligned sentences from languages that are supported by the model, where most of the languages are tail languages (275). The retrieval task is performed without any training: we directly use the model after continued pretraining to encode all sentences. Each sentence is represented by taking the average of the contextual embedding at the \textbf{8th} layer. We then compute the top-10 accuracy for each pair (English and another language) by finding the nearest neighbors (in the other language) of the representation of each English sentence.

\paragraph{SR-T.} We use up to 1000 English-aligned sentences from Tatoeba, which mainly contains head languages (70). The evaluation setting is the same as SR-B and top-10 accuracy is reported. 

\paragraph{Taxi1500.} We finetune the continued pretrained model (a sequence-level classification model in 6 classes) on the English train set and select the best checkpoint using the English dev set. We train each model for a maximum of 40 epochs with early stopping on a single GTX 1080 Ti GPU. Adam optimizer is used, the learning rate is set to 1e-5 and the effective batch size is set to 16 (batch size of 8 and gradient accumulation of 2). We then evaluate the zero-shot performance by evaluating the finetuned model on the test sets of all other language-scripts. F1 score is reported for each language-script.

\paragraph{NER.} We finetune the continued pretrained model (a token-level classification model in 7 classes) on the English train set and select the best checkpoint using the English dev set. We train each model for a maximum of 10 epochs with early stopping on a single GTX 1080 Ti GPU. Adam optimizer is used, the learning rate is set to 2e-5 and the effective batch size is set to 32 (batch size of 8 and gradient accumulation of 4). We then evaluate the zero-shot performance by evaluating the finetuned model on the test sets of all other language-scripts. F1 score is reported for each language-script.

\paragraph{POS.} We finetune the continued pretrained model (a token-level classification model in 18 classes) on the English train set and select the best checkpoint using the English dev set. We train each model for a maximum of 10 epochs with early stopping on a single GTX 1080 Ti GPU. Adam optimizer is used, the learning rate is set to 2e-5 and the effective batch size is set to 32 (batch size of 8 and gradient accumulation of 4). We then evaluate the zero-shot performance by evaluating the finetuned model on the test sets of all other language-scripts. F1 score is reported for each language-script.

\section{Multilingual Word Vectors and Coverage}\seclabel{coverage}
Two important factors that influence the effectiveness of \frameworkname initialization are (1) the quality of the external multilingual word vectors and (2) the coverage of the multilingual word vectors in terms of languages and new subwords in the target model. 

In this work, we use \embname \citep{liu2023crosslingual}, multilingual word vectors learned from colexification\footnote{Colexifications are a linguistic phenomenon where different meanings are expressed by the same word.} \citep{franccois2008semantic} graphs built from 1,335 translations (one for a specific language identified by its ISO-639-3 code) of Parallel Bible Corpus \citep{mayer-cysouw-2014-creating}. The patterns of colexifications are extracted by Conceptualizer \citep{liu-etal-2023-crosslingual}, a statistic concept-grams alignment method. The tokens in the word vectors are ngrams (mostly word types as the algorithm prefers longer ngrams) within whitespace tokenized words. According to \citet{liu2023crosslingual}, \embname outperforms a bunch of strong multilingual word vector baselines on crosslingual transfer tasks, especially for low-resource languages. we therefore choose to use \embname as our multilingual word vectors.

\begin{table}
\setlength{\belowcaptionskip}{-0.4cm}
\centering
\small
\setlength\tabcolsep{5pt}
\begin{tabular}{lrrrr}
        \toprule
        Source models & Copy & Similarity & Random & Coverage\\
        \midrule
        \multirow{1}{*}{RoBERTa} 
        & 27K & 179K & 195K & 51.5\%\\
        \multirow{1}{*}{XLM-R} 
        & 255K & 84K & 62K & 84.6\% \\
        \bottomrule
    \end{tabular}
\caption{\label{tab:coverage}
The number of subwords being initialized by copying from the original embeddings (\textbf{Copy}); through the similarity-based method introduced in \frameworkname (\textbf{Similarity}); and randomly from a Gaussian distribution (\textbf{Random}) when using \embname as the external multilingual word vectors. Coverage shows the percentage of the subword being wisely initialized: (Copy + Similarity) / (Copy + Similarity + Random). The coverage is high for both of the source models. As the new vocabulary is extended from XLM-R, many subword embeddings are directly copied when using XLM-R as the source model.
}
\end{table}

We want as many as possible subwords to be initialized wisely (either directly copied for shared subwords or initialized by the similarity-based method in \frameworkname), instead of being randomly initialized from a Gaussian distribution. This requires that the chosen external multilingual word vectors cover many subwords. Therefore we report the number of subwords being initialized (1) \textbf{by copying}, (2) \textbf{through the similarity-based method}, and (3) \textbf{randomly} when using \embname as our external multilingual word vectors in Table \ref{tab:coverage}. We see that for either the monolingual model as the source model (RoBERTa) or the multilingual model as the source model (XLM-R), the coverage (subwords being wisely initialized over all subwords) is more than 50\%, indicating that the words included in \embname cover a large number of subwords even though it is trained from a genre-specific corpus.

\section{Redundancy in Multilingual PLMs}\seclabel{redundancy}

\begin{figure}
  \centering
  \includegraphics[width=0.5\textwidth]{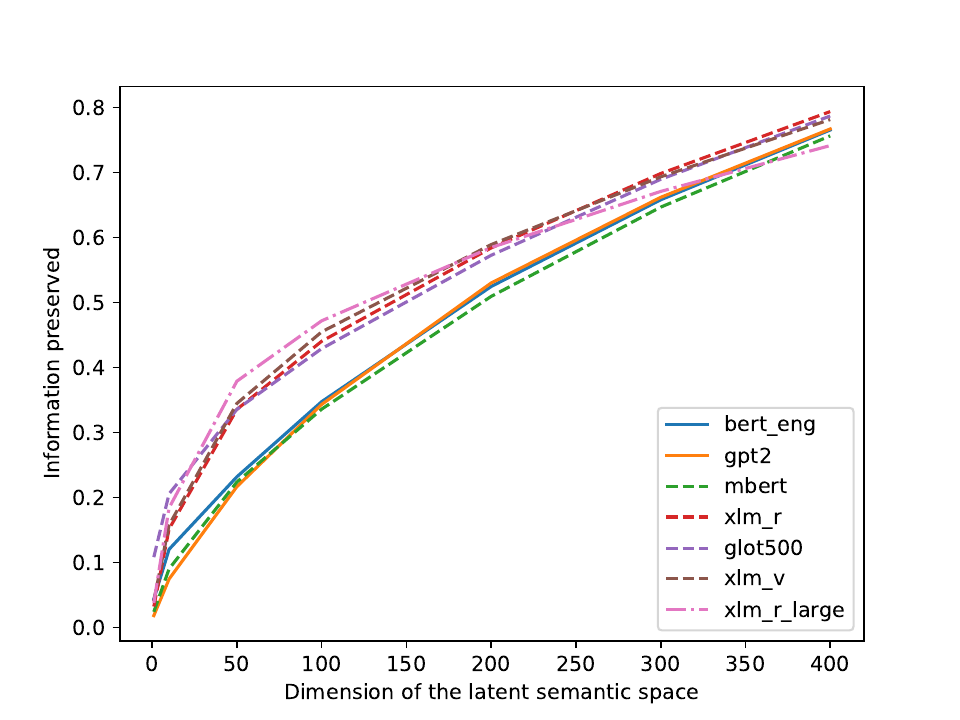}
  \caption{Information preserved (percentage of variance explained by the selected components) under different dimensions of the semantic space (number of principal components). Generally trend: multilingual models generally preserve more information than monolingual ones when embeddings are reduced to the same dimension.}
  \label{fig:redundancy}
\end{figure}

To figure out how ``redundant'' the embeddings are in monolingual or multilingual PLMs, we use principle component analysis (PCA) to perform dimension reduction to the embeddings of various PLMs. We select monolingual PLMs: BERT \citep{devlin-etal-2019-bert} of English and GPT-2 \citep{radford2019language}, and multilingual PLMs: mBERT \cite{devlin-etal-2019-bert}, base and large versions of XLM-R \citep{conneau-etal-2020-unsupervised},  Glot500-m \citep{imani-etal-2023-glot500} and XLM-V \citep{liang-etal-2023-xlm}. The embedding dimension and vocabulary size of each PLM are shown in Table \ref{tab:additional_model_size}. We report how much variance is explained (information preserved) when keeping different numbers of principle components in the sorted order by their eigenvalues (until the first 400 components) in Figure \ref{fig:redundancy}. The general trend is that multilingual PLMs tend to be more ``redundant'' than monolingual ones: only keeping the first 100 components, about 50\% variance can be explained in Glot500-m and XLM-R-large embeddings. Similarly, the information preserved is more than 40\% in XLM-R-base and XLM-V, which is higher than the percentage in monolingual models GPT-2 and English BERT (about 30\% is preserved), when the first 100 components are kept. 

\begin{table}
    \setlength{\belowcaptionskip}{-0.3cm}
    \small
    \centering
    \def\tablesep{0.2cm}
    \begin{tabular}{
      @{\hspace{\tablesep}}l@{\hspace{\tablesep}}
      @{\hspace{\tablesep}}r@{\hspace{\tablesep}}
      @{\hspace{\tablesep}}r@{\hspace{\tablesep}}
    }
    \toprule
    PLM & emb dim. & |V| \\
    \midrule
    BERT-eng  & 768  & 31K \\
    % BERT-zho  & 768  & 21K \\
    GPT-2  & 768  & 50K \\
    \midrule
    mBERT & 768 & 120K\\
    XLM-R-base & 768 & 250K\\
    XLM-R-large & 1024 & 250K\\
    Glot500-m & 768 & 401K\\
    XLM-V & 768 & 901K\\
    \bottomrule
    \end{tabular}
    \caption{Embedding dimensions and vocabulary size of several monolingual and multilingual PLMs.}
    \label{tab:additional_model_size}
\end{table}

% \begin{table*}[h]
%   \centering
%   \small
%   \begin{tabular}{l|r|rr|rr}
% \toprule
%  & |L| & rand is better & one of OFA-mono is better & rand is better & one of OFA-multi is better \\
% \midrule
% SR-B & 369 & 0 & \textbf{369} & 23 & \textbf{346} \\
% SR-T & 98 & 1 & \textbf{97} & 24 & \textbf{74} \\
% Taxi1500 & 351 & 5 & \textbf{346} & 31 & \textbf{320} \\
% NER & 164 & 10 & \textbf{154} & 27 & \textbf{137} \\
% POS & 91 & 4 & \textbf{87} & 12 & \textbf{79} \\
% \bottomrule
%   \end{tabular}
%   \caption{Number of languages in each downstream task that benefits from \frameworkname framework compared with randomly initializing the new subwords. |L| is the total number of languages for each task.
%   \label{tab:whosbetter}}
% \end{table*}

We also assume this ``redundancy'' is related to the crosslinguality of the PLMs. If the embedding matrix is more redundant, this indicates the many tokens referring to the same concept from different languages share similar representation space, therefore better crosslinguality is expected. For example, both base and large versions of XLM-R are more redundant than mBERT according to Figure \ref{fig:redundancy}, indicating better crosslinguality, which aligns with the finding that XLM-R constantly outperforms mBERT in many NLP downstream tasks \citep{conneau-etal-2020-unsupervised}. However, the high redundancy, in turn, suggests an unnecessary over-parameterization. Thus we could use matrix factorization to remove some redundancy to reduce the number of parameters while not sacrificing much performance, which is exactly what we propose in the \frameworkname framework: replacing the cumbersome embedding matrix with two smaller matrices.

\section{Fine-grained Aggregated Results for Each Task}\seclabel{complete_results_family_script}

To better illustrate how \frameworkname can influence the continued pretraining and thus influence the crosslinguality, we additionally report the aggregated results for 7 major language families in Table \ref{tab:family_sr_b}, \ref{tab:family_sr_t}, \ref{tab:family_taxi1500}, \ref{tab:family_ner}, \ref{tab:family_pos} and 5 major script groups in Table \ref{tab:script_sr_b}, \ref{tab:script_sr_t}, \ref{tab:script_taxi1500}, \ref{tab:script_ner}, \ref{tab:script_pos} for each task. It is clear that the models continued pretrained with \frameworkname show better performance for each language family and script group in every downstream task. We also show the number of languages that benefit from \frameworkname in each downstream task in Table \ref{tab:whosbetter}.

\begin{table*}[ht]
    \scriptsize
    \centering
    \setlength{\tabcolsep}{1.0mm}{}
    \begin{tabular}{lrrrrrrrrr}
\toprule
 & (indo1319, 54) & (atla1278, 2) & (aust1307, 7) & (turk1311, 7) & (sino1245, 3) & (maya1287, 0) & (afro1255, 5) & (other, 20) & (all, 98) \\
\midrule
RoBERTa & 6.9 & 12.6 & 3.2 & 3.0 & 1.3 & - & 2.6 & 5.5 & 5.8 \\
RoBERTa-rand & 26.5 & 24.7 & 22.0 & 16.2 & 11.6 & - & 9.8 & 16.3 & 22.0 \\
\frameworkname-mono-100 & 31.4 & 23.9 & 23.2 & 18.0 & 25.9 & - & 13.3 & 21.6 & 26.6 \\
\frameworkname-mono-200 & \underline{39.3} & 32.4 & \underline{31.2} & \underline{25.2} & \underline{38.1} & - & \underline{16.3} & \underline{27.5} & \underline{33.9} \\
\frameworkname-mono-400 & \textbf{52.1} & \textbf{34.0} & \textbf{39.4} & \textbf{37.2} & \textbf{59.6} & - & \textbf{28.0} & \textbf{41.9} & \textbf{46.7} \\
\frameworkname-mono-768 & 33.3 & \underline{33.9} & 29.3 & 21.9 & 26.4 & - & 11.7 & 21.7 & 28.6 \\
\midrule
XLM-R & 63.4 & 29.6 & 35.3 & 41.6 & 62.0 & - & 41.4 & 56.7 & 56.6 \\
XLM-R-rand & \underline{70.7} & \textbf{49.4} & \underline{48.4} & \textbf{64.6} & \textbf{76.2} & - & \textbf{46.1} & \underline{63.6} & \underline{65.7} \\
\frameworkname-multi-100 & 66.5 & 46.6 & 46.3 & 58.4 & 69.0 & - & \underline{44.5} & 57.7 & 61.3 \\
\frameworkname-multi-200 & 63.8 & 45.4 & 44.9 & 54.7 & 65.5 & - & 40.5 & 54.9 & 58.5 \\
\frameworkname-multi-400 & 67.2 & 42.8 & 46.7 & 58.5 & 68.9 & - & 44.4 & 59.0 & 61.8 \\
\frameworkname-multi-768 & \textbf{71.7} & \underline{48.1} & \textbf{49.6} & \underline{63.9} & \underline{75.4} & - & 44.3 & \textbf{64.3} & \textbf{66.3} \\
        \bottomrule
    \end{tabular}
    \caption{Aggregated performance of the models for 7 major language families on \textbf{SR-T}. We report the average performance for \textbf{indo1319} (Indo-European), \textbf{atla1278 } (Atlantic-Congo), \textbf{aust1307} (Austronesian), \textbf{turk1311} (Turkic), \textbf{sino1245} (Sino-Tibetan), \textbf{maya1287} (Mayan), and \textbf{afro1255} (Afro-Asiatic). We classify the remaining languages into the group ``\textbf{other}''. In addition, we report the average over all languages (group ``\textbf{all}'').  The number of languages is shown in the parentheses.
    \textbf{Bold} (\underline{underlined}): best (second-best) result for each task.}
    \label{tab:family_sr_t}
\end{table*}

\begin{table*}[ht]
    \scriptsize
    \centering
    \setlength{\tabcolsep}{1.0mm}{}
    \begin{tabular}{lrrrrrrrrr}
\toprule
 & (indo1319, 87) & (atla1278, 68) & (aust1307, 51) & (turk1311, 18) & (sino1245, 22) & (maya1287, 15) & (afro1255, 11) & (other, 79) & (all, 351) \\
\midrule
RoBERTa & 8.2 & 4.9 & 5.2 & 4.9 & 4.8 & 4.9 & 5.2 & 5.2 & 5.8 \\
RoBERTa-rand & 23.4 & 13.5 & 17.4 & 10.6 & 11.9 & 9.7 & 10.8 & 11.0 & 15.5 \\
\frameworkname-mono-100 & 27.4 & 13.6 & 20.0 & 18.2 & 16.5 & 11.6 & 10.7 & 13.0 & 18.1 \\
\frameworkname-mono-200 & \underline{40.9} & \underline{26.8} & \underline{34.7} & \underline{33.1} & \underline{25.6} & \underline{26.0} & \underline{21.1} & \underline{27.3} & \underline{31.6} \\
\frameworkname-mono-400 & \textbf{50.5} & \textbf{32.1} & \textbf{38.2} & \textbf{42.3} & \textbf{33.3} & \textbf{26.4} & \textbf{25.3} & \textbf{33.3} & \textbf{37.9} \\
\frameworkname-mono-768 & 32.5 & 21.5 & 26.1 & 29.3 & 14.7 & 13.8 & 14.2 & 19.5 & 23.8 \\
\midrule
XLM-R & 48.4 & 13.3 & 23.4 & 30.9 & 21.9 & 11.1 & 19.3 & 20.9 & 26.7 \\
XLM-R-rand & 61.0 & 42.8 & 52.9 & 59.9 & 48.6 & 40.6 & 37.9 & 45.0 & 50.3 \\
\frameworkname-multi-100 & 59.3 & 47.2 & 54.5 & 60.7 & 53.0 & 45.4 & 37.4 & 47.9 & 52.1 \\
\frameworkname-multi-200 & 57.0 & 45.3 & 53.2 & 57.3 & 52.0 & \underline{45.7} & 36.0 & 47.0 & 50.5 \\
\frameworkname-multi-400 & \underline{61.7} & \underline{49.1} & \underline{55.7} & \textbf{65.3} & \underline{54.2} & \underline{45.7} & \underline{38.6} & \underline{48.9} & \underline{53.8} \\
\frameworkname-multi-768 & \textbf{64.7} & \textbf{51.4} & \textbf{58.4} & \underline{65.2} & \textbf{56.2} & \textbf{49.6} & \textbf{43.4} & \textbf{53.0} & \textbf{56.7} \\
        \bottomrule
    \end{tabular}
    \caption{Aggregated performance of the models for 7 major language families on \textbf{Taxi1500}. We report the average performance for \textbf{indo1319} (Indo-European), \textbf{atla1278 } (Atlantic-Congo), \textbf{aust1307} (Austronesian), \textbf{turk1311} (Turkic), \textbf{sino1245} (Sino-Tibetan), \textbf{maya1287} (Mayan), and \textbf{afro1255} (Afro-Asiatic). We classify the remaining languages into the group ``\textbf{other}''. In addition, we report the average over all languages (group ``\textbf{all}'').  The number of languages is shown in the parentheses.
    \textbf{Bold} (\underline{underlined}): best (second-best) result for each task.}
    \label{tab:family_taxi1500}
\end{table*}

\begin{table*}[ht]
    \scriptsize
    \centering
    \setlength{\tabcolsep}{1.0mm}{}
    \begin{tabular}{lrrrrrrrrr}
\toprule
 & (indo1319, 94) & (atla1278, 5) & (aust1307, 12) & (turk1311, 12) & (sino1245, 7) & (maya1287, 0) & (afro1255, 6) & (other, 28) & (all, 164) \\
\midrule
RoBERTa & 31.2 & 39.4 & 41.1 & 17.4 & 6.9 & - & 12.8 & 23.8 & 28.2 \\
RoBERTa-rand & 56.3 & 53.4 & 53.2 & 48.1 & 25.3 & - & 35.8 & 41.8 & 50.8 \\
\frameworkname-mono-100 & 57.9 & 52.0 & \underline{54.7} & 48.4 & 30.9 & - & 41.3 & 45.5 & 52.9 \\
\frameworkname-mono-200 & \underline{60.8} & 50.3 & 52.6 & 54.9 & \textbf{34.8} & - & \underline{46.2} & \underline{50.1} & \underline{56.0} \\
\frameworkname-mono-400 & \textbf{63.6} & \textbf{57.6} & \textbf{55.7} & \textbf{58.0} & \underline{34.1} & - & \textbf{49.2} & \textbf{51.9} & \textbf{58.6} \\
\frameworkname-mono-768 & 60.2 & \underline{55.4} & 54.4 & \underline{55.3} & 28.7 & - & 40.5 & 47.5 & 55.1 \\
\midrule
XLM-R & 61.0 & 46.5 & 49.7 & 50.7 & 26.4 & - & 47.5 & 50.9 & 55.3 \\
XLM-R-rand & 66.1 & 56.9 & \textbf{60.2} & 60.8 & 35.0 & - & 52.2 & 55.8 & 61.4 \\
\frameworkname-multi-100 & 63.9 & 56.2 & 56.3 & 59.3 & 32.2 & - & 53.3 & 53.3 & 59.2 \\
\frameworkname-multi-200 & 65.1 & \underline{61.5} & 56.7 & 61.1 & 36.9 & - & 50.9 & 54.7 & 60.6 \\
\frameworkname-multi-400 & \textbf{67.8} & \textbf{63.8} & \underline{59.0} & \textbf{64.4} & \textbf{40.5} & - & \textbf{56.3} & \underline{56.6} & \textbf{63.3} \\
\frameworkname-multi-768 & \underline{66.9} & 61.2 & 58.5 & \underline{61.7} & \underline{38.1} & - & \underline{54.8} & \textbf{56.9} & \underline{62.4} \\
        \bottomrule
    \end{tabular}
    \caption{Aggregated performance of the models for 7 major language families on \textbf{NER}. We report the average performance for \textbf{indo1319} (Indo-European), \textbf{atla1278 } (Atlantic-Congo), \textbf{aust1307} (Austronesian), \textbf{turk1311} (Turkic), \textbf{sino1245} (Sino-Tibetan), \textbf{maya1287} (Mayan), and \textbf{afro1255} (Afro-Asiatic). We classify the remaining languages into the group ``\textbf{other}''. In addition, we report the average over all languages (group ``\textbf{all}'').  The number of languages is shown in the parentheses.
    \textbf{Bold} (\underline{underlined}): best (second-best) result for each task.}
    \label{tab:family_ner}
\end{table*}

\begin{table*}[ht]
    \scriptsize
    \centering
    \setlength{\tabcolsep}{1.0mm}{}
    \begin{tabular}{lrrrrrrrrr}
\toprule
 & (indo1319, 54) & (atla1278, 2) & (aust1307, 4) & (turk1311, 5) & (sino1245, 3) & (maya1287, 1) & (afro1255, 6) & (other, 16) & (all, 91) \\
\midrule
RoBERTa & 29.1 & 21.2 & 33.1 & 24.0 & 11.7 & 26.5 & 15.6 & 23.1 & 26.3 \\
RoBERTa-rand & 66.0 & 48.8 & 63.9 & 44.1 & 18.6 & 54.7 & 43.0 & 42.6 & 57.0 \\
\frameworkname-mono-100 & 68.8 & 50.1 & 68.0 & 52.6 & 16.1 & 62.3 & 49.4 & 46.2 & 60.4 \\
\frameworkname-mono-200 & 68.5 & 47.4 & 67.3 & 53.8 & \underline{23.6} & 55.7 & 50.5 & \underline{48.6} & 60.8 \\
\frameworkname-mono-400 & \textbf{73.3} & \textbf{62.0} & \textbf{72.9} & \textbf{65.5} & 21.3 & \textbf{63.9} & \textbf{59.9} & \textbf{53.5} & \textbf{66.4} \\
\frameworkname-mono-768 & \underline{69.8} & \underline{53.9} & \underline{70.3} & \underline{55.2} & \textbf{26.1} & \textbf{63.9} & \underline{50.9} & 47.9 & \underline{62.1} \\
\midrule
XLM-R & 75.4 & 24.1 & 70.1 & 57.3 & 22.2 & 28.7 & 54.0 & 54.1 & 65.6 \\
XLM-R-rand & 76.9 & \textbf{62.4} & 74.3 & 70.7 & 28.9 & \textbf{62.7} & 63.5 & 59.6 & 70.5 \\
\frameworkname-multi-100 & 76.6 & 60.0 & 72.4 & 70.8 & 33.0 & 57.5 & 61.9 & 57.5 & 69.7 \\
\frameworkname-multi-200 & 76.9 & \underline{62.3} & 73.4 & 71.2 & 33.7 & \underline{60.6} & 64.5 & 58.6 & 70.5 \\
\frameworkname-multi-400 & \textbf{77.9} & 60.8 & \textbf{75.2} & \textbf{73.6} & \underline{33.8} & 59.0 & \textbf{66.8} & \underline{59.8} & \underline{71.6} \\
\frameworkname-multi-768 & \underline{77.7} & 60.3 & \underline{75.1} & \underline{72.6} & \textbf{36.0} & 60.1 & \underline{66.4} & \textbf{61.1} & \textbf{71.7} \\
        \bottomrule
    \end{tabular}
    \caption{Aggregated performance of the models for 7 major language families on \textbf{POS}. We report the average performance for \textbf{indo1319} (Indo-European), \textbf{atla1278 } (Atlantic-Congo), \textbf{aust1307} (Austronesian), \textbf{turk1311} (Turkic), \textbf{sino1245} (Sino-Tibetan), \textbf{maya1287} (Mayan), and \textbf{afro1255} (Afro-Asiatic). We classify the remaining languages into the group ``\textbf{other}''. In addition, we report the average over all languages (group ``\textbf{all}'').  The number of languages is shown in the parentheses.
    \textbf{Bold} (\underline{underlined}): best (second-best) result for each task.}
    \label{tab:family_pos}
\end{table*}

\begin{table*}[ht]
    \footnotesize
    \centering
    \setlength{\tabcolsep}{1.0mm}{}
    \begin{tabular}{lrrrrrrr}
\toprule
 & (Latn, 290) & (Cyrl, 28) & (Hani, 4) & (Arab, 11) & (Deva, 8) & (other, 28) & (all, 369) \\
\midrule
RoBERTa & 3.7 & 2.1 & 2.6 & 2.2 & 2.1 & 2.1 & 3.4 \\
RoBERTa-rand & 12.7 & 11.7 & 12.1 & 10.0 & 9.2 & 6.0 & 11.9 \\
\frameworkname-mono-100 & 15.0 & 16.8 & 17.8 & 15.2 & 16.3 & 11.1 & 14.9 \\
\frameworkname-mono-200 & \underline{18.1} & \underline{23.2} & \underline{25.7} & \underline{21.8} & \underline{22.2} & \underline{15.5} & \underline{18.6} \\
\frameworkname-mono-400 & \textbf{27.9} & \textbf{37.6} & \textbf{36.4} & \textbf{36.9} & \textbf{39.6} & \textbf{28.0} & \textbf{29.2} \\
\frameworkname-mono-768 & \underline{18.1} & 20.8 & 24.4 & 19.4 & 19.8 & 10.9 & 17.9 \\
\midrule
XLM-R & 16.2 & 25.5 & 30.4 & 36.3 & 32.1 & 33.8 & 19.3 \\
XLM-R-rand & 41.9 & 59.2 & 40.9 & 50.8 & 57.4 & 46.3 & 44.2 \\
\frameworkname-multi-100 & 35.8 & 51.8 & 37.1 & 42.9 & 46.8 & 33.2 & 37.3 \\
\frameworkname-multi-200 & 41.8 & 60.6 & 40.6 & 51.2 & 56.1 & 42.9 & 43.9 \\
\frameworkname-multi-400 & \underline{46.4} & \textbf{64.5} & \textbf{41.9} & \textbf{54.7} & \textbf{61.6} & \underline{48.5} & \underline{48.5} \\
\frameworkname-multi-768 & \textbf{46.8} & \underline{63.5} & \underline{41.3} & \underline{53.6} & \underline{61.3} & \textbf{48.9} & \textbf{48.7} \\
        \bottomrule
    \end{tabular}
    \caption{Aggregated performance of the models for 5 major script groups on \textbf{SR-B}. We report the average performance for \textbf{Latn} (Latin), \textbf{Cyrl} (Cyrillic), \textbf{Hani} (Hani), \textbf{Arab} (Arabic), and \textbf{Deva} (Devanagari). We classify the remaining languages into the group ``\textbf{other}''. In addition, we report the average over all languages (group ``\textbf{all}'').  The number of languages is shown in the parentheses.
    \textbf{Bold} (\underline{underlined}): best (second-best) result for each task.}
    \label{tab:script_sr_b}
\end{table*}

\begin{table*}[ht]
    \footnotesize
    \centering
    \setlength{\tabcolsep}{1.0mm}{}
    \begin{tabular}{lrrrrrrr}
\toprule
 & (Latn, 64) & (Cyrl, 10) & (Hani, 3) & (Arab, 5) & (Deva, 2) & (other, 14) & (all, 98) \\
\midrule
RoBERTa & 7.9 & 2.0 & 1.3 & 1.2 & 1.2 & 2.1 & 5.8 \\
RoBERTa-rand & 27.5 & 17.0 & 11.6 & 11.2 & 12.2 & 7.9 & 22.0 \\
\frameworkname-mono-100 & 30.0 & 23.8 & 25.9 & 17.1 & 19.4 & 17.5 & 26.6 \\
\frameworkname-mono-200 & \underline{37.2} & \underline{32.2} & \underline{38.1} & \underline{26.4} & \underline{29.2} & \underline{23.0} & \underline{33.9} \\
\frameworkname-mono-400 & \textbf{47.2} & \textbf{48.4} & \textbf{59.6} & \textbf{44.0} & \textbf{53.4} & \textbf{40.4} & \textbf{46.7} \\
\frameworkname-mono-768 & 33.3 & 26.0 & 26.4 & 18.1 & 20.8 & 13.8 & 28.6 \\
\midrule
XLM-R & 55.7 & 55.5 & 62.0 & 53.6 & 68.6 & 59.7 & 56.6 \\
XLM-R-rand & \underline{64.8} & \underline{68.7} & \textbf{76.2} & \textbf{66.2} & \textbf{76.9} & \underline{63.2} & \underline{65.7} \\
\frameworkname-multi-100 & 61.4 & 63.0 & 69.0 & 60.9 & 67.4 & 57.3 & 61.3 \\
\frameworkname-multi-200 & 58.7 & 60.1 & 65.5 & 56.3 & 64.4 & 54.4 & 58.5 \\
\frameworkname-multi-400 & 61.5 & 63.4 & 68.9 & 60.8 & 68.0 & 60.2 & 61.8 \\
\frameworkname-multi-768 & \textbf{65.9} & \textbf{69.4} & \underline{75.4} & \underline{63.8} & \underline{76.2} & \textbf{63.3} & \textbf{66.3} \\
        \bottomrule
    \end{tabular}
    \caption{Aggregated performance of the models for 5 major script groups on \textbf{SR-T}. We report the average performance for \textbf{Latn} (Latin), \textbf{Cyrl} (Cyrillic), \textbf{Hani} (Hani), \textbf{Arab} (Arabic), and \textbf{Deva} (Devanagari). We classify the remaining languages into the group ``\textbf{other}''. In addition, we report the average over all languages (group ``\textbf{all}'').  The number of languages is shown in the parentheses.
    \textbf{Bold} (\underline{underlined}): best (second-best) result for each task.}
    \label{tab:script_sr_t}
\end{table*}

\begin{table*}[ht]
    \footnotesize
    \centering
    \setlength{\tabcolsep}{1.0mm}{}
    \begin{tabular}{lrrrrrrr}
\toprule
 & (Latn, 281) & (Cyrl, 25) & (Hani, 4) & (Arab, 8) & (Deva, 7) & (other, 26) & (all, 351) \\
\midrule
RoBERTa & 6.1 & 4.9 & 4.9 & 4.9 & 4.9 & 5.0 & 5.8 \\
RoBERTa-rand & 16.0 & 12.6 & 26.0 & 20.4 & 16.2 & 9.3 & 15.5 \\
\frameworkname-mono-100 & 17.7 & 20.5 & 28.1 & 25.2 & 16.0 & 16.4 & 18.1 \\
\frameworkname-mono-200 & \underline{30.4} & \underline{34.5} & \underline{49.3} & \underline{37.6} & \underline{41.2} & \underline{35.1} & \underline{31.6} \\
\frameworkname-mono-400 & \textbf{35.8} & \textbf{40.9} & \textbf{60.9} & \textbf{46.4} & \textbf{57.3} & \textbf{47.1} & \textbf{37.9} \\
\frameworkname-mono-768 & 22.8 & 28.4 & 31.0 & 32.9 & 37.3 & 23.1 & 23.8 \\
\midrule
XLM-R & 22.5 & 30.2 & \underline{66.6} & 48.5 & 49.5 & 49.5 & 26.7 \\
XLM-R-rand & 48.3 & 55.8 & 64.8 & 60.4 & 64.1 & 57.2 & 50.3 \\
\frameworkname-multi-100 & 50.5 & 56.6 & 66.3 & 58.5 & 66.4 & 56.7 & 52.1 \\
\frameworkname-multi-200 & 49.2 & 53.9 & 63.6 & 58.8 & 62.3 & 53.7 & 50.5 \\
\frameworkname-multi-400 & \underline{52.0} & \underline{60.9} & 64.5 & \textbf{61.2} & \underline{68.2} & \underline{59.2} & \underline{53.8} \\
\frameworkname-multi-768 & \textbf{55.1} & \textbf{62.1} & \textbf{67.3} & \underline{61.0} & \textbf{70.7} & \textbf{62.9} & \textbf{56.7} \\
        \bottomrule
    \end{tabular}
    \caption{Aggregated performance of the models for 5 major script groups on \textbf{Taxi1500}. We report the average performance for \textbf{Latn} (Latin), \textbf{Cyrl} (Cyrillic), \textbf{Hani} (Hani), \textbf{Arab} (Arabic), and \textbf{Deva} (Devanagari). We classify the remaining languages into the group ``\textbf{other}''. In addition, we report the average over all languages (group ``\textbf{all}'').  The number of languages is shown in the parentheses.
    \textbf{Bold} (\underline{underlined}): best (second-best) result for each task.}
    \label{tab:script_taxi1500}
\end{table*}

\begin{table*}[ht]
    \footnotesize
    \centering
    \setlength{\tabcolsep}{1.0mm}{}
    \begin{tabular}{lrrrrrrr}
\toprule
 & (Latn, 104) & (Cyrl, 17) & (Hani, 4) & (Arab, 10) & (Deva, 5) & (other, 24) & (all, 164) \\
\midrule
RoBERTa & 42.5 & 4.4 & 1.4 & 2.4 & 2.8 & 3.5 & 28.2 \\
RoBERTa-rand & 60.7 & 49.8 & 17.1 & 30.3 & 37.7 & 25.9 & 50.8 \\
\frameworkname-mono-100 & 59.8 & 52.6 & \textbf{24.5} & 38.5 & 42.0 & 36.5 & 52.9 \\
\frameworkname-mono-200 & 61.2 & \textbf{60.6} & \textbf{24.5} & \underline{45.6} & 47.2 & \underline{41.5} & \underline{56.0} \\
\frameworkname-mono-400 & \textbf{64.3} & \underline{59.5} & 21.7 & \textbf{48.8} & \textbf{55.0} & \textbf{44.4} & \textbf{58.6} \\
\frameworkname-mono-768 & \underline{62.3} & 55.6 & 22.9 & 40.2 & \underline{48.6} & 36.3 & 55.1 \\
\midrule
XLM-R & 60.3 & 51.8 & 23.1 & 45.0 & 56.9 & 45.2 & 55.3 \\
XLM-R-rand & 66.9 & 64.3 & 21.9 & 51.3 & 56.4 & 47.4 & 61.4 \\
\frameworkname-multi-100 & 64.6 & 63.0 & 19.8 & 49.4 & 54.0 & 45.1 & 59.2 \\
\frameworkname-multi-200 & 65.6 & 63.6 & 20.5 & 52.6 & 55.2 & 47.7 & 60.6 \\
\frameworkname-multi-400 & \textbf{68.2} & \textbf{65.0} & \textbf{29.5} & \textbf{55.9} & \textbf{57.8} & \textbf{50.6} & \textbf{63.3} \\
\frameworkname-multi-768 & \underline{67.2} & \textbf{65.0} & \underline{23.8} & \underline{53.6} & \textbf{57.8} & \textbf{50.6} & \underline{62.4} \\
        \bottomrule
    \end{tabular}
    \caption{Aggregated performance of the models for 5 major script groups on \textbf{NER}. We report the average performance for \textbf{Latn} (Latin), \textbf{Cyrl} (Cyrillic), \textbf{Hani} (Hani), \textbf{Arab} (Arabic), and \textbf{Deva} (Devanagari). We classify the remaining languages into the group ``\textbf{other}''. In addition, we report the average over all languages (group ``\textbf{all}'').  The number of languages is shown in the parentheses.
    \textbf{Bold} (\underline{underlined}): best (second-best) result for each task.}
    \label{tab:script_ner}
\end{table*}

\begin{table*}[ht]
    \footnotesize
    \centering
    \setlength{\tabcolsep}{1.0mm}{}
    \begin{tabular}{lrrrrrrr}
\toprule
 & (Latn, 57) & (Cyrl, 8) & (Hani, 3) & (Arab, 5) & (Deva, 3) & (other, 15) & (all, 91) \\
\midrule
RoBERTa & 31.8 & 24.8 & 11.7 & 14.1 & 2.4 & 18.0 & 26.3 \\
RoBERTa-rand & 64.7 & 62.5 & 18.6 & 44.8 & 31.5 & 41.9 & 57.0 \\
\frameworkname-mono-100 & 65.9 & 67.9 & 16.1 & 53.4 & 40.7 & 50.6 & 60.4 \\
\frameworkname-mono-200 & 65.7 & 68.5 & \underline{23.6} & 54.8 & \underline{41.0} & \underline{51.8} & 60.8 \\
\frameworkname-mono-400 & \textbf{70.7} & \textbf{76.4} & 21.3 & \textbf{61.7} & \textbf{44.5} & \textbf{60.0} & \textbf{66.4} \\
\frameworkname-mono-768 & \underline{67.7} & \underline{69.9} & \textbf{26.1} & \underline{55.0} & 40.9 & 50.3 & \underline{62.1} \\
\midrule
XLM-R & 68.1 & 66.7 & 22.2 & 65.8 & 58.3 & 65.9 & 65.6 \\
XLM-R-rand & 73.8 & 78.3 & 28.9 & 66.1 & 57.1 & 66.0 & 70.5 \\
\frameworkname-multi-100 & 72.8 & 78.3 & 33.0 & 67.4 & 54.4 & 64.9 & 69.7 \\
\frameworkname-multi-200 & 73.2 & \underline{79.7} & 33.7 & \underline{67.8} & 56.0 & 66.6 & 70.5 \\
\frameworkname-multi-400 & \underline{74.5} & \textbf{80.4} & \underline{33.8} & \textbf{69.3} & \textbf{59.5} & \underline{66.9} & \underline{71.6} \\
\frameworkname-multi-768 & \textbf{74.7} & 79.5 & \textbf{36.0} & 67.7 & \underline{58.4} & \textbf{67.2} & \textbf{71.7} \\
        \bottomrule
    \end{tabular}
    \caption{Aggregated performance of the models for 5 major script groups on \textbf{POS}. We report the average performance for \textbf{Latn} (Latin), \textbf{Cyrl} (Cyrillic), \textbf{Hani} (Hani), \textbf{Arab} (Arabic), and \textbf{Deva} (Devanagari). We classify the remaining languages into the group ``\textbf{other}''. In addition, we report the average over all languages (group ``\textbf{all}'').  The number of languages is shown in the parentheses.
    \textbf{Bold} (\underline{underlined}): best (second-best) result for each task.}
    \label{tab:script_pos}
\end{table*}

\section{Complete Results for Each Task and Language}\seclabel{complete_results}

We report the complete results for all tasks and languages in Table \ref{tab:srb_table1}, \ref{tab:srb_table2}, \ref{tab:srb_table3} \ref{tab:srb_table4} (SR-B), Table \ref{tab:srt_table} (SR-T), Table \ref{tab:taxi1500_table1}, \ref{tab:taxi1500_table2}, \ref{tab:taxi1500_table3}, \ref{tab:taxi1500_table4} (Taxi1500), Table \ref{tab:ner_table1}, \ref{tab:ner_table2} (NER), and Table \ref{tab:pos_table} (POS).

\begin{table*}
\centering
\resizebox{\textwidth}{!}{
\begin{tabular}{l|rrrrrr|rrrrrr}
    \toprule
    Language-script & RoBERTa & RoBERTa-rand & \frameworkname-mono-100 & \frameworkname-mono-200 & \frameworkname-mono-400 & \frameworkname-mono-768 & XLM-R & XLM-R-rand & \frameworkname-multi-100 & \frameworkname-multi-200 & \frameworkname-multi-400 & \frameworkname-multi-768 \\
    \midrule
    ace\_Latn & 2.6 & 10.4 & 18.8 & 18.0 & \textbf{24.4} & 16.4 & 4.4 & 51.8 & 39.4 & 43.2 & 48.4 & \textbf{54.6} \\
ach\_Latn & 4.0 & 10.6 & 8.4 & 9.8 & \textbf{18.0} & 13.0 & 4.4 & 46.6 & 30.4 & 37.2 & \textbf{54.4} & 43.2 \\
acr\_Latn & 2.0 & 8.2 & 9.8 & 10.0 & \textbf{15.6} & 9.4 & 2.6 & 18.4 & 13.2 & 19.8 & \textbf{27.0} & 22.8 \\
afr\_Latn & 4.4 & 14.8 & 22.8 & 31.8 & \textbf{33.8} & 23.8 & \textbf{76.8} & 71.4 & 65.0 & 71.6 & 71.0 & 74.4 \\
agw\_Latn & 3.8 & 10.4 & 14.8 & 15.4 & \textbf{25.4} & 20.8 & 5.8 & 31.4 & 25.8 & 31.0 & \textbf{41.6} & 37.8 \\
ahk\_Latn & 2.4 & 3.2 & 3.4 & 2.8 & \textbf{4.0} & 3.8 & 3.0 & 3.2 & 2.6 & 3.6 & \textbf{4.4} & 3.2 \\
aka\_Latn & 5.0 & 10.8 & 14.0 & 18.8 & \textbf{32.6} & 18.6 & 5.0 & 46.4 & 46.0 & 48.2 & \textbf{53.0} & 51.6 \\
\bottomrule
    \end{tabular}
}
    \caption{Top-10 accuracy of baselines and models initialized with \frameworkname on \textbf{SR-B} (Part I).}\label{tab:srb_table1}
\end{table*}

\begin{table*}
\centering
\resizebox{\textwidth}{!}{
    \begin{tabular}{l|rrrrrr|rrrrrr}
    \toprule
    Language-script & RoBERTa & RoBERTa-rand & \frameworkname-mono-100 & \frameworkname-mono-200 & \frameworkname-mono-400 & \frameworkname-mono-768 & XLM-R & XLM-R-rand & \frameworkname-multi-100 & \frameworkname-multi-200 & \frameworkname-multi-400 & \frameworkname-multi-768 \\
    \midrule
    gla\_Latn & 3.6 & 7.4 & 8.2 & 9.4 & \textbf{14.8} & 9.2 & 25.2 & 38.2 & 27.4 & 34.8 & \textbf{41.4} & 39.8 \\
gle\_Latn & 3.0 & 6.6 & 9.6 & 11.2 & \textbf{16.0} & 10.4 & 35.0 & \textbf{41.6} & 29.0 & 35.4 & 37.0 & 40.8 \\
glv\_Latn & 3.0 & 8.4 & 9.8 & 11.8 & \textbf{13.8} & 10.4 & 5.8 & 35.2 & 31.4 & 39.8 & \textbf{46.6} & 44.8 \\
gom\_Latn & 3.6 & 5.4 & 5.6 & 5.6 & \textbf{13.0} & 8.0 & 6.0 & 37.2 & 27.0 & 37.6 & \textbf{47.6} & 42.0 \\
gor\_Latn & 3.0 & 8.4 & 8.8 & 10.2 & \textbf{12.8} & 9.8 & 3.8 & 20.0 & 16.6 & 21.2 & \textbf{27.6} & 25.0 \\
grc\_Grek & 2.2 & 3.6 & 9.8 & 11.2 & \textbf{21.8} & 11.0 & 17.4 & 49.6 & 33.2 & 41.2 & 46.8 & \textbf{50.4} \\
guc\_Latn & 2.4 & 6.0 & 5.4 & 5.2 & \textbf{8.8} & 6.4 & 3.4 & 9.0 & 9.4 & 9.6 & 10.8 & \textbf{11.2} \\
\bottomrule
    \end{tabular}
}
    \caption{Top-10 accuracy of baselines and models initialized with \frameworkname on \textbf{SR-B} (Part II).}\label{tab:srb_table2}
\end{table*}

\begin{table*}
\centering
\resizebox{\textwidth}{!}{
    \begin{tabular}{l|rrrrrr|rrrrrr}
    \toprule
    Language-script & RoBERTa & RoBERTa-rand & \frameworkname-mono-100 & \frameworkname-mono-200 & \frameworkname-mono-400 & \frameworkname-mono-768 & XLM-R & XLM-R-rand & \frameworkname-multi-100 & \frameworkname-multi-200 & \frameworkname-multi-400 & \frameworkname-multi-768 \\
    \midrule
    mam\_Latn & 2.6 & 5.8 & 6.0 & 6.4 & \textbf{7.4} & 6.0 & 3.8 & 11.0 & 8.4 & 11.6 & \textbf{13.2} & 10.4 \\
mar\_Deva & 2.2 & 9.4 & 20.0 & 24.6 & \textbf{43.2} & 17.8 & 66.2 & 65.2 & 54.4 & 60.6 & 66.2 & \textbf{72.8} \\
mau\_Latn & 2.0 & 3.4 & 2.8 & 3.2 & \textbf{4.6} & 4.2 & 2.4 & 3.2 & 3.2 & 3.6 & 3.2 & \textbf{3.8} \\
mbb\_Latn & 2.4 & 11.8 & 10.0 & 16.0 & \textbf{26.8} & 17.0 & 3.0 & 25.0 & 19.4 & 27.6 & \textbf{34.6} & 30.8 \\
mck\_Latn & 2.6 & 13.0 & 19.6 & 20.0 & \textbf{35.2} & 22.2 & 5.2 & 65.6 & 53.0 & 59.0 & 64.8 & \textbf{67.2} \\
mcn\_Latn & 3.4 & 10.0 & 9.6 & 12.0 & \textbf{23.4} & 12.0 & 6.0 & 45.4 & 25.0 & 36.2 & 43.0 & \textbf{46.0} \\
mco\_Latn & 2.2 & 5.0 & 4.6 & 4.4 & \textbf{6.0} & 4.8 & 2.6 & \textbf{8.0} & 4.8 & 6.8 & 7.8 & 6.6 \\
\bottomrule
    \end{tabular}
}
    \caption{Top-10 accuracy of baselines and models initialized with \frameworkname on \textbf{SR-B} (Part III).}\label{tab:srb_table3}
\end{table*}

\begin{table*}
\centering
\resizebox{\textwidth}{!}{
    \begin{tabular}{l|rrrrrr|rrrrrr}
    \toprule
    Language-script & RoBERTa & RoBERTa-rand & \frameworkname-mono-100 & \frameworkname-mono-200 & \frameworkname-mono-400 & \frameworkname-mono-768 & XLM-R & XLM-R-rand & \frameworkname-multi-100 & \frameworkname-multi-200 & \frameworkname-multi-400 & \frameworkname-multi-768 \\
    \midrule
    sah\_Cyrl & 1.6 & 8.6 & 13.2 & 16.6 & \textbf{29.0} & 16.2 & 6.2 & 49.8 & 37.4 & 47.0 & \textbf{52.6} & 52.0 \\
san\_Deva & 1.6 & 5.4 & 6.0 & 8.0 & \textbf{14.2} & 7.2 & 13.8 & 23.2 & 16.4 & 19.6 & \textbf{27.2} & 24.8 \\
san\_Latn & 2.4 & 3.2 & 3.0 & 3.0 & \textbf{5.2} & 3.0 & 4.6 & 9.4 & 4.4 & 6.4 & \textbf{10.0} & 9.8 \\
sba\_Latn & 2.4 & 8.4 & 11.6 & 14.4 & \textbf{15.8} & 10.2 & 2.8 & 26.0 & 23.0 & 21.8 & 30.2 & \textbf{34.0} \\
seh\_Latn & 3.0 & 15.4 & 22.4 & 27.2 & \textbf{40.0} & 29.2 & 6.4 & 63.6 & 62.4 & 75.2 & \textbf{75.6} & 74.2 \\
sin\_Sinh & 1.8 & 4.6 & 7.4 & 10.6 & \textbf{19.0} & 8.2 & 44.8 & 48.0 & 28.4 & 37.8 & \textbf{48.4} & 44.4 \\
slk\_Latn & 3.8 & 11.4 & 21.0 & 20.0 & \textbf{37.2} & 21.8 & \textbf{75.2} & 66.4 & 54.8 & 63.4 & 64.0 & 68.8 \\
\bottomrule
    \end{tabular}
}
    \caption{Top-10 accuracy of baselines and models initialized with \frameworkname on \textbf{SR-B} (Part IV).}\label{tab:srb_table4}
\end{table*}
\begin{table*}
\centering
\resizebox{\textwidth}{!}{
    \begin{tabular}{l|rrrrrr|rrrrrr}
    \toprule
    Language-script & RoBERTa & RoBERTa-rand & \frameworkname-mono-100 & \frameworkname-mono-200 & \frameworkname-mono-400 & \frameworkname-mono-768 & XLM-R & XLM-R-rand & \frameworkname-multi-100 & \frameworkname-multi-200 & \frameworkname-multi-400 & \frameworkname-multi-768 \\
    \midrule
    afr\_Latn & 4.1 & 19.9 & 31.5 & 33.7 & \textbf{40.0} & 22.9 & 71.9 & 74.2 & 73.7 & 65.0 & 69.1 & \textbf{75.6} \\
amh\_Ethi & 6.5 & 14.3 & 16.1 & 16.7 & \textbf{28.6} & 16.7 & 35.1 & 42.9 & 40.5 & 39.3 & \textbf{46.4} & 44.0 \\
ara\_Arab & 1.0 & 8.4 & 13.0 & 18.8 & \textbf{31.4} & 11.5 & \textbf{59.2} & 57.3 & 56.8 & 51.9 & 50.3 & 54.2 \\
arz\_Arab & 2.1 & 13.6 & 18.2 & 23.1 & \textbf{39.4} & 17.0 & 32.5 & \textbf{57.4} & 56.2 & 47.8 & 52.0 & 49.3 \\
ast\_Latn & 42.5 & 63.8 & 61.4 & 63.0 & \textbf{70.1} & 60.6 & 59.8 & \textbf{83.5} & 81.1 & 75.6 & 81.9 & \textbf{83.5} \\
aze\_Latn & 2.2 & 22.2 & 24.6 & 37.4 & \textbf{50.8} & 31.7 & 62.6 & \textbf{75.7} & 68.3 & 66.0 & 69.8 & 73.9 \\
bel\_Cyrl & 1.4 & 15.9 & 29.4 & 34.4 & \textbf{54.6} & 25.9 & 70.0 & 75.0 & 70.4 & 67.4 & 69.5 & \textbf{75.3} \\
\bottomrule
    \end{tabular}
}
    \caption{Top-10 accuracy of baselines and models initialized with \frameworkname on \textbf{SR-T}.}\label{tab:srt_table}
\end{table*}
\begin{table*}
\centering
\resizebox{\textwidth}{!}{
    \begin{tabular}{l|rrrrrr|rrrrrr}
    \toprule
    Language-script & RoBERTa & RoBERTa-rand & \frameworkname-mono-100 & \frameworkname-mono-200 & \frameworkname-mono-400 & \frameworkname-mono-768 & XLM-R & XLM-R-rand & \frameworkname-multi-100 & \frameworkname-multi-200 & \frameworkname-multi-400 & \frameworkname-multi-768 \\
    \midrule
    ace\_Latn & 6.1 & 17.7 & 26.7 & 38.8 & \textbf{50.4} & 33.6 & 13.4 & \textbf{66.6} & 62.2 & 62.4 & 63.9 & 65.7 \\
ach\_Latn & 4.9 & 9.5 & 8.7 & 18.1 & \textbf{28.3} & 18.6 & 10.9 & 33.4 & \textbf{46.7} & 43.0 & 38.9 & 41.9 \\
acr\_Latn & 4.9 & 5.3 & 14.2 & 24.1 & \textbf{36.2} & 15.4 & 8.8 & 44.3 & \textbf{53.0} & 46.6 & 52.5 & 45.6 \\
afr\_Latn & 8.6 & 41.7 & 41.6 & 43.3 & \textbf{44.4} & 37.4 & \textbf{65.7} & 61.8 & 59.3 & 62.8 & 65.5 & 58.7 \\
agw\_Latn & 7.3 & 8.5 & 13.5 & 34.6 & \textbf{38.3} & 30.1 & 13.9 & 48.7 & 59.8 & \textbf{60.3} & 49.7 & 58.3 \\
ahk\_Latn & 4.9 & 4.9 & \textbf{6.7} & 4.9 & 4.9 & 4.9 & 9.3 & \textbf{14.2} & 6.1 & 4.9 & 4.9 & 9.1 \\
aka\_Latn & 4.9 & 18.9 & 20.2 & 27.2 & 32.8 & \textbf{35.7} & 9.1 & 39.2 & \textbf{57.4} & 53.9 & 51.3 & 43.4 \\
\bottomrule
    \end{tabular}
}
    \caption{F1 scores of baselines and models initialized with \frameworkname on \textbf{Taxi1500} (Part I).}\label{tab:taxi1500_table1}
\end{table*}

\begin{table*}
\centering
\resizebox{\textwidth}{!}{
    \begin{tabular}{l|rrrrrr|rrrrrr}
    \toprule
    Language-script & RoBERTa & RoBERTa-rand & \frameworkname-mono-100 & \frameworkname-mono-200 & \frameworkname-mono-400 & \frameworkname-mono-768 & XLM-R & XLM-R-rand & \frameworkname-multi-100 & \frameworkname-multi-200 & \frameworkname-multi-400 & \frameworkname-multi-768 \\
    \midrule
    gle\_Latn & 4.9 & 12.2 & 5.2 & 18.0 & 20.6 & \textbf{24.1} & 40.1 & 38.6 & 37.8 & 34.7 & 37.5 & \textbf{54.6} \\
glv\_Latn & 4.9 & 23.8 & 12.1 & 8.3 & \textbf{32.9} & 15.4 & 11.7 & 43.8 & \textbf{45.0} & 41.8 & 43.0 & 42.8 \\
gom\_Latn & 4.9 & 11.4 & \textbf{14.9} & 8.0 & 10.3 & 7.6 & 13.0 & 31.7 & \textbf{48.3} & 37.3 & 47.2 & 44.8 \\
gor\_Latn & 4.9 & 7.3 & 18.2 & 19.4 & 28.0 & \textbf{29.0} & 18.5 & 47.5 & 41.9 & 49.3 & 45.6 & \textbf{54.0} \\
guc\_Latn & 4.9 & 20.3 & 19.2 & 25.0 & \textbf{36.5} & 19.8 & 8.7 & 35.2 & 26.9 & 32.9 & 33.9 & \textbf{41.7} \\
gug\_Latn & 4.9 & 15.4 & 16.7 & \textbf{29.3} & 26.3 & 24.1 & 15.3 & 39.0 & \textbf{51.6} & 44.8 & 50.0 & 44.7 \\
guj\_Gujr & 4.9 & 8.4 & 25.3 & 43.3 & \textbf{51.9} & 30.2 & 62.9 & 72.4 & 68.4 & 69.6 & 68.1 & \textbf{73.7} \\
\bottomrule
    \end{tabular}
}
    \caption{F1 scores of baselines and models initialized with \frameworkname on \textbf{Taxi1500} (Part II).}\label{tab:taxi1500_table2}
\end{table*}

\begin{table*}
\centering
\resizebox{\textwidth}{!}{
    \begin{tabular}{l|rrrrrr|rrrrrr}
    \toprule
    Language-script & RoBERTa & RoBERTa-rand & \frameworkname-mono-100 & \frameworkname-mono-200 & \frameworkname-mono-400 & \frameworkname-mono-768 & XLM-R & XLM-R-rand & \frameworkname-multi-100 & \frameworkname-multi-200 & \frameworkname-multi-400 & \frameworkname-multi-768 \\
    \midrule
    mar\_Deva & 4.9 & 10.9 & 14.1 & 38.9 & \textbf{41.4} & 33.7 & 60.7 & 56.8 & 68.8 & 61.4 & 67.6 & \textbf{74.4} \\
mau\_Latn & \textbf{4.9} & \textbf{4.9} & \textbf{4.9} & \textbf{4.9} & \textbf{4.9} & \textbf{4.9} & 6.5 & 7.1 & \textbf{8.2} & 4.9 & 4.9 & 7.1 \\
mbb\_Latn & 4.9 & 19.3 & 19.1 & 33.3 & 34.8 & \textbf{39.0} & 8.7 & 57.4 & \textbf{61.5} & 55.8 & 58.7 & 58.8 \\
mck\_Latn & 4.9 & 20.6 & 21.1 & 25.1 & \textbf{36.8} & 18.8 & 18.2 & 44.4 & 49.2 & 43.1 & 51.0 & \textbf{51.1} \\
mcn\_Latn & 4.9 & 9.8 & 4.9 & 21.3 & \textbf{22.1} & 15.0 & 10.7 & \textbf{49.1} & 47.1 & 44.5 & 41.9 & 47.5 \\
mco\_Latn & 4.9 & 4.9 & 4.9 & 7.3 & \textbf{17.6} & 10.5 & 8.2 & 14.4 & \textbf{35.0} & 21.5 & 23.3 & 18.8 \\
mdy\_Ethi & 4.9 & 4.9 & 13.4 & 27.0 & \textbf{37.3} & 7.0 & 4.9 & 48.7 & 56.7 & 54.9 & 55.8 & \textbf{58.8} \\
\bottomrule
    \end{tabular}
}
    \caption{F1 scores of baselines and models initialized with \frameworkname on \textbf{Taxi1500} (Part III).}\label{tab:taxi1500_table3}
\end{table*}

\begin{table*}
\centering
\resizebox{\textwidth}{!}{
    \begin{tabular}{l|rrrrrr|rrrrrr}
    \toprule
    Language-script & RoBERTa & RoBERTa-rand & \frameworkname-mono-100 & \frameworkname-mono-200 & \frameworkname-mono-400 & \frameworkname-mono-768 & XLM-R & XLM-R-rand & \frameworkname-multi-100 & \frameworkname-multi-200 & \frameworkname-multi-400 & \frameworkname-multi-768 \\
    \midrule
    rus\_Cyrl & 4.9 & 19.5 & 47.2 & 50.0 & \textbf{64.3} & 44.3 & 68.9 & 69.5 & 66.3 & 64.8 & 71.9 & \textbf{75.0} \\
sag\_Latn & 4.7 & 17.1 & 4.9 & 21.6 & \textbf{29.1} & 7.2 & 11.9 & 38.4 & 41.1 & \textbf{52.4} & 42.9 & 48.4 \\
sah\_Cyrl & 4.9 & 4.9 & 26.7 & \textbf{28.5} & 28.3 & 24.6 & 14.8 & 53.4 & 50.6 & 57.8 & \textbf{64.6} & 60.8 \\
sba\_Latn & 4.9 & 13.8 & 5.0 & \textbf{29.1} & 25.3 & 23.8 & 7.9 & 43.0 & 40.2 & \textbf{45.9} & 40.7 & 38.9 \\
seh\_Latn & 4.9 & 19.8 & 17.2 & 27.9 & \textbf{36.0} & 28.8 & 13.4 & 43.8 & \textbf{62.7} & 48.3 & 55.1 & 51.0 \\
sin\_Sinh & 4.9 & 10.0 & 16.4 & 42.6 & \textbf{47.3} & 26.0 & 65.8 & 69.5 & 60.2 & 59.2 & 58.9 & \textbf{72.2} \\
slk\_Latn & 4.9 & 20.8 & 29.5 & 33.3 & \textbf{40.1} & 28.5 & \textbf{72.6} & 57.7 & 49.6 & 45.3 & 54.0 & 63.5 \\
\bottomrule
    \end{tabular}
}
    \caption{F1 scores of baselines and models initialized with \frameworkname on \textbf{Taxi1500} (Part IV).}\label{tab:taxi1500_table4}
\end{table*}
\begin{table*}
\centering
\resizebox{\textwidth}{!}{
    \begin{tabular}{l|rrrrrr|rrrrrr}
    \toprule
    Language-script & RoBERTa & RoBERTa-rand & \frameworkname-mono-100 & \frameworkname-mono-200 & \frameworkname-mono-400 & \frameworkname-mono-768 & XLM-R & XLM-R-rand & \frameworkname-multi-100 & \frameworkname-multi-200 & \frameworkname-multi-400 & \frameworkname-multi-768 \\
    \midrule
    ace\_Latn & 27.5 & 36.3 & 39.6 & 36.1 & \textbf{41.1} & 40.4 & 33.7 & 42.4 & 42.5 & \textbf{42.8} & 40.1 & 40.0 \\
afr\_Latn & 47.0 & 72.6 & 68.4 & 72.8 & \textbf{75.1} & 73.6 & 75.7 & \textbf{76.7} & 74.9 & 73.5 & 76.6 & 75.7 \\
als\_Latn & 43.1 & 76.1 & 75.1 & 76.9 & 80.3 & \textbf{80.6} & 61.8 & 81.5 & 78.7 & 81.4 & \textbf{82.7} & 80.8 \\
amh\_Ethi & 0.0 & 36.6 & 37.8 & \textbf{42.9} & 42.1 & 29.3 & 41.8 & 30.9 & 45.5 & 36.4 & \textbf{54.0} & 44.1 \\
ara\_Arab & 2.6 & 25.1 & 35.8 & 39.1 & \textbf{40.9} & 37.9 & 45.4 & 50.4 & 48.9 & 51.6 & 50.1 & \textbf{51.7} \\
arg\_Latn & 59.8 & 76.4 & 74.5 & 77.9 & \textbf{79.9} & 75.9 & 73.7 & \textbf{82.3} & 76.6 & 80.1 & 77.3 & 78.4 \\
arz\_Arab & 2.1 & 35.2 & 42.4 & \textbf{51.0} & 49.4 & 43.8 & 48.0 & 53.8 & \textbf{58.7} & 53.0 & 55.7 & 55.2 \\
\bottomrule
    \end{tabular}
}
    \caption{F1 scores of baselines and models initialized with \frameworkname on \textbf{NER} (Part I).}\label{tab:ner_table1}
\end{table*}

\begin{table*}
\centering
\resizebox{\textwidth}{!}{
    \begin{tabular}{l|rrrrrr|rrrrrr}
    \toprule
    Language-script & RoBERTa & RoBERTa-rand & \frameworkname-mono-100 & \frameworkname-mono-200 & \frameworkname-mono-400 & \frameworkname-mono-768 & XLM-R & XLM-R-rand & \frameworkname-multi-100 & \frameworkname-multi-200 & \frameworkname-multi-400 & \frameworkname-multi-768 \\
    \midrule
    lij\_Latn & 24.6 & 41.4 & 36.8 & 40.0 & \textbf{45.5} & 39.9 & 38.7 & 43.4 & 40.7 & 42.3 & \textbf{47.8} & 41.2 \\
lim\_Latn & 42.4 & 66.7 & 64.9 & \textbf{68.1} & 63.8 & 66.7 & 62.6 & 67.3 & 71.4 & 71.1 & 67.0 & \textbf{72.3} \\
lin\_Latn & 33.7 & 51.1 & 39.0 & 41.9 & \textbf{52.3} & 46.7 & 37.1 & 52.1 & 53.3 & 53.9 & \textbf{60.7} & 53.7 \\
lit\_Latn & 41.2 & 64.9 & 61.7 & 63.8 & \textbf{66.5} & 64.1 & 71.9 & 72.8 & 70.6 & 71.4 & 73.0 & \textbf{73.8} \\
lmo\_Latn & 38.6 & 68.9 & 65.1 & 66.8 & 70.3 & \textbf{71.2} & 67.3 & 74.6 & 70.4 & \textbf{75.6} & 75.4 & 71.0 \\
ltz\_Latn & 36.5 & 63.8 & 61.2 & 66.5 & \textbf{68.8} & 68.2 & 49.0 & \textbf{69.1} & 64.6 & 66.6 & 68.6 & 68.0 \\
lzh\_Hani & 1.1 & 9.2 & \textbf{14.6} & 14.4 & 12.2 & 12.9 & 15.7 & 11.9 & 15.3 & 10.7 & \textbf{21.2} & 9.5 \\
\bottomrule
    \end{tabular}
}
    \caption{F1 scores of baselines and models initialized with \frameworkname on \textbf{NER} (Part II).}\label{tab:ner_table2}
\end{table*}
\begin{table*}
\centering
\resizebox{\textwidth}{!}{
    \begin{tabular}{l|rrrrrr|rrrrrr}
    \toprule
    Language-script & RoBERTa & RoBERTa-rand & \frameworkname-mono-100 & \frameworkname-mono-200 & \frameworkname-mono-400 & \frameworkname-mono-768 & XLM-R & XLM-R-rand & \frameworkname-multi-100 & \frameworkname-multi-200 & \frameworkname-multi-400 & \frameworkname-multi-768 \\
    \midrule
    afr\_Latn & 29.7 & 79.9 & 80.5 & 81.5 & \textbf{83.7} & 82.4 & \textbf{89.3} & 88.1 & 87.3 & 87.3 & 84.6 & 88.5 \\
ajp\_Arab & 20.8 & 50.5 & 52.5 & 55.3 & \textbf{63.6} & 54.9 & 63.0 & 68.0 & 67.9 & 67.4 & \textbf{72.0} & 71.7 \\
aln\_Latn & 16.0 & 39.0 & 40.0 & 39.3 & \textbf{46.9} & 42.1 & \textbf{54.1} & 48.8 & 49.6 & 50.2 & 50.6 & 51.9 \\
amh\_Ethi & 5.7 & 47.5 & 43.0 & 47.7 & \textbf{58.6} & 47.0 & 63.9 & 63.2 & 58.9 & 63.8 & \textbf{65.8} & \textbf{65.8} \\
ara\_Arab & 16.5 & 53.5 & 55.8 & 57.0 & \textbf{63.8} & 61.8 & \textbf{67.9} & 65.6 & 65.6 & 66.9 & 67.3 & 66.9 \\
bam\_Latn & 23.3 & \textbf{40.6} & 38.4 & 33.1 & 33.5 & 37.2 & 25.1 & \textbf{44.0} & 32.2 & 35.0 & 37.9 & 40.5 \\
bel\_Cyrl & 23.5 & 77.3 & 81.3 & 80.6 & \textbf{84.6} & 81.2 & 86.0 & 84.7 & 85.2 & \textbf{86.4} & 84.8 & 86.0 \\
\bottomrule
    \end{tabular}
}
    \caption{F1 scores of baselines and models initialized with \frameworkname on \textbf{POS}.}\label{tab:pos_table}
\end{table*}

% This is a section in the appendix.

\end{document}